\pdfoutput=1
\documentclass[10pt,twocolumn,letterpaper]{article}

\usepackage{iccv}
\usepackage{times}
\usepackage{epsfig}
\usepackage{graphicx}
\usepackage{amsmath}
\usepackage{amssymb}
\usepackage{array}
\usepackage{bm}
\usepackage{multirow}
\usepackage{booktabs}
\usepackage[table,xcdraw]{xcolor}
\usepackage{amssymb}
\usepackage[ruled,linesnumbered]{algorithm2e}
\SetKwInput{KwInput}{Input}                
\SetKwInput{KwOutput}{Output}
\usepackage{docmute}

\usepackage[breaklinks=true,bookmarks=false]{hyperref}

\iccvfinalcopy 

\def\iccvPaperID{3871} 

\ificcvfinal\fi

\begin{document}

\title{Deep Metric Learning for Open World Semantic Segmentation}

\author{Jun Cen \quad Peng Yun \quad Junhao Cai \quad Michael Yu Wang \quad Ming Liu\\
The Hong Kong University of Science and Technology\\
{\tt\small \{jcenaa, pyun, jcaiaq\}@connect.ust.hk, \{mywang, eelium\}@ust.hk}
}

\maketitle
\ificcvfinal\thispagestyle{empty}\fi

\begin{abstract}
   Classical close-set semantic segmentation networks have limited ability to detect out-of-distribution (OOD) objects, which is important for safety-critical applications such as autonomous driving. Incrementally learning these OOD objects with few annotations is an ideal way to enlarge the knowledge base of the deep learning models. In this paper, we propose an \textbf{open world semantic segmentation system} that includes two modules: (1) an open-set semantic segmentation module to detect both in-distribution and OOD objects. (2) an incremental few-shot learning module to gradually incorporate those OOD objects into its existing knowledge base. This open world semantic segmentation system behaves like a human being, which is able to identify OOD objects and gradually learn them with corresponding supervision. We adopt the Deep Metric Learning Network (DMLNet) with contrastive clustering to implement open-set semantic segmentation. Compared to other open-set semantic segmentation methods, our DMLNet achieves state-of-the-art performance on three challenging open-set semantic segmentation datasets without using additional data or generative models. On this basis, two incremental few-shot learning methods are further proposed to progressively improve the DMLNet with the annotations of OOD objects.
\end{abstract}
\vspace{-0.7cm}

\section{Introduction}

\begin{figure}[t]
\begin{center}
   \includegraphics[width=1\linewidth]{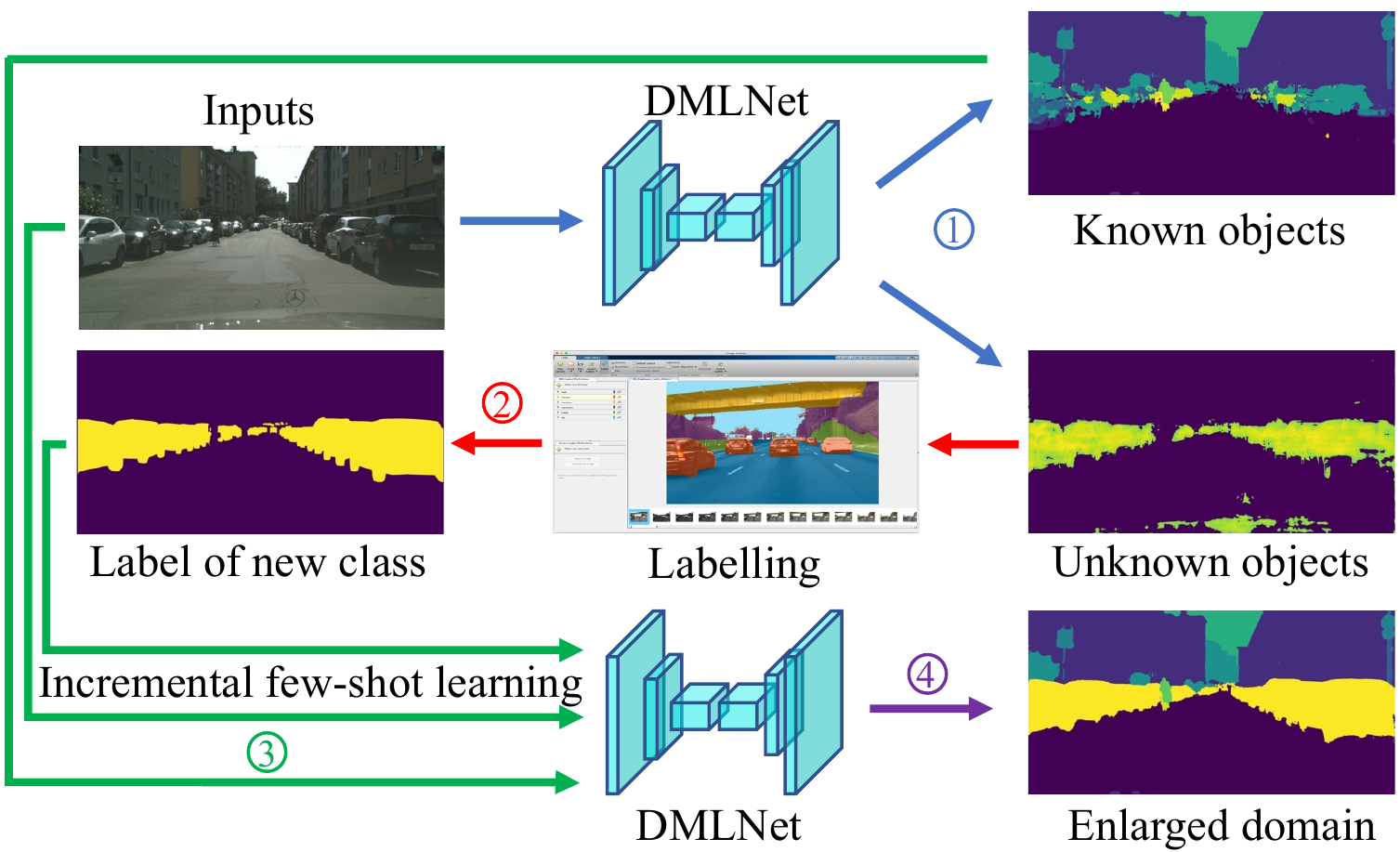}
\end{center}
   \caption{{\bf Open world semantic segmentation system.} Step \textbf{1}: Identify both known and unknown objects (blue arrow). Step \textbf{2}: Annotate for unknown objects (red arrow). Step \textbf{3}: Apply incremental few-shot learning to increase the classification range of the network (green arrow). Step \textbf{4}: After incremental few-shot learning, DMLNet can output the result in a larger domain (purple arrow).}
\label{fig:head}
\vspace{-0.5cm}
\end{figure}

Deep convolutional networks have achieved tremendous success in semantic segmentation tasks~\cite{chen2018encoder,Zhao2016}, benefiting from high-quality datasets~\cite{cordts2016cityscapes,huang2018apolloscape,Neuhold2017}. These semantic segmentation networks are used as the perception system in many applications, like autonomous driving~\cite{janai2020computer}, medical diagnose~\cite{challen2019artificial}, etc. However, most of these perception systems are \textit{close-set} and \textit{static}. Close-set semantic segmentation is subject to the assumption that all classes in testing are already involved during training, which is not true in the open world. A close-set system could cause disastrous consequences in security-critical applications, like autonomous driving, if it wrongly assigns an in-distribution label to an OOD object~\cite{Bozhinoski2019}. Meanwhile, a static perception system cannot update its knowledge base according to what it has seen, and therefore, it is limited to particular scenarios and needs to be retrained after a certain amout of time. To solve these problems, we propose an \textit{open-set} and \textit{dynamic} perception system called the open world semantic segmentation system. It contains two modules: (1) an \textbf{open-set semantic segmentation module} to detect OOD objects and assign correct labels to in-distribution objects. (2) an \textbf{incremental few-shot learning module} to incorporate those unknown objects progressively into its existing knowledge base. The whole pipeline of our proposed open world semantic segmentation system is shown in Fig.~\ref{fig:head}.

Both open-set semantic segmentation and incremental few-shot learning are not well solved. For open-set semantic segmentation, the essential part is to identify OOD pixels among all pixels in one image which is called \textit{anomaly segmentation}. The typical approach of anomaly segmentation is to adapt methods of image-level open-set classification for pixel-level open-set classification. These methods include uncertainty estimation-based methods~\cite{hendrycks2016baseline,gal2016dropout,hendrycks2019scaling,DBLP:conf/nips/Lakshminarayanan17} and autoencoder-based methods~\cite{baur2018deep,creusot2015real}. However, both of these methods have been proved ineffective in driving scenarios as uncertainty estimation-based methods give many false-positive outlier detections~\cite{bevandic2021dense} and autoencoders cannot regenerate complicated urban scenarios~\cite{Lis2019}. Recently, generative adversarial network-based (GAN-based) methods~\cite{Lis2019, xia2020synthesize} have been proved effective but they are far from lightweight as they need several deep networks in the pipeline. For incremental few-shot learning, we have to deal with challenges not only from the incremental learning, such as catastrophic forgetting~\cite{mccloskey1989catastrophic}, but also from the few-shot learning, including extracting representative features from a small number of samples~\cite{snell2017prototypical}.

\begin{figure}[h]
\begin{center}
  \includegraphics[width=1\linewidth]{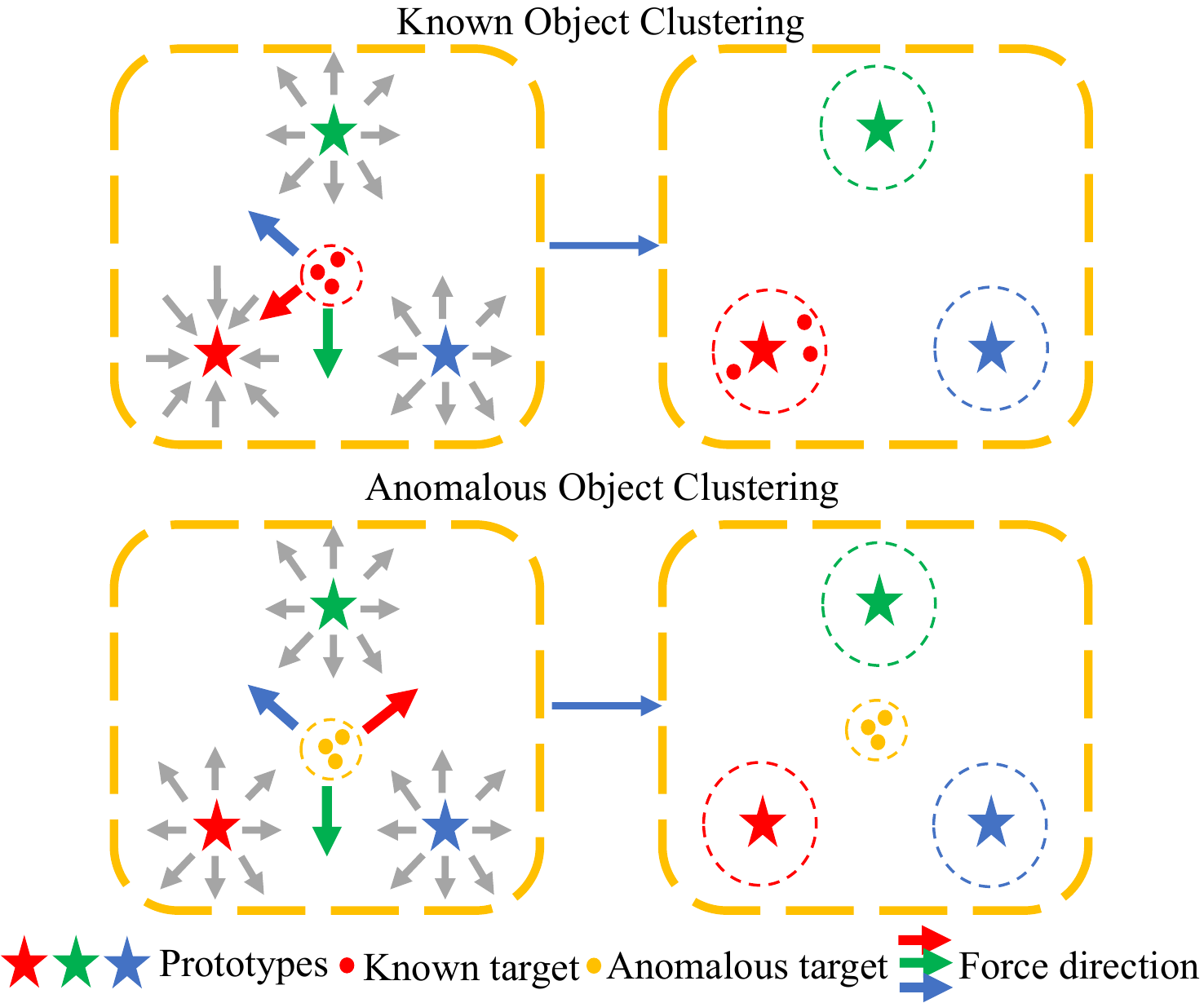}
    \end{center}
\vspace{-0.3cm}
  \caption{{\bf Contrastive clustering of DMLNet.} During inference, known objects will be attracted by the prototype of the same class and repelled by remaining prototypes. In the end they will aggregate around the particular prototype. In contrast, anomalous objects will be repelled by all prototypes so they will aggregate at the middle of the metric space. }
\label{fig:contrastive}
\vspace{-0.3cm}
\end{figure}

In this paper, we propose to use the DMLNet to solve the open world semantic segmentation problem. The reasons are threefold: (1) The classification principle of the DMLNet is based on contrastive clustering, which is effective to identify anomalous objects, as shown in Fig.~\ref{fig:contrastive}. (2) The DMLNet combined with prototypes is very suitable for few-shot tasks~\cite{snell2017prototypical}. (3) Incremental learning of the DMLNet can be implemented by adding new prototypes, which is a natural and useful approach~\cite{DBLP:conf/cvpr/RebuffiKSL17}. Based on the DMLNet architecture, we develop two unknown identification criteria for the open-set semantic segmentation module and two methods for the incremental few-shot learning module. Both modules are verified to be effective and lightweight according to our experiments. To summarize, our contributions are the following:
\vspace{-0.2cm}
\begin{itemize}
\setlength{\itemsep}{0pt}
\setlength{\parsep}{0pt}
\setlength{\parskip}{0pt}
    \item We are the first to introduce the open world semantic segmentation system, which is more robust and practical in real-world applications.
    \item Our proposed open-set semantic segmentation module based on the DMLNet achieves state-of-the-art performance on three challenging datasets.
    \item Our proposed methods for the few-shot incremental learning module alleviate the catastrophic forgetting problem to a large extent.
    \item An open world semantic segmentation system is fulfilled by combining our proposed open-set semantic segmentation module and incremental few-shot learning module. 
\end{itemize}

\section{Related Work}
\label{sec:related work}

\subsection{Anomaly semantic segmentation}
\label{sec:realted work 1}
The approaches of anomaly semantic segmentation can be divided into two trends: uncertainty estimation-based methods and generative model-based methods. The baseline of uncertainty estimation is the maximum softmax probability (MSP), which was first proposed in~\cite{hendrycks2016baseline}. Instead of using the softmax probability, Dan \textit{et al.}~\cite{hendrycks2019scaling} proposed to use the maximum logit (MaxLogit) and achieved better anomaly segmentation performance.  Bayesian networks adopt a probabilistic view of deep learning networks, so their weights and outputs are probability distributions instead of specific numbers~\cite{mackay1995bayesian, DBLP:conf/nips/KendallG17}. In practice, Dropout~\cite{gal2016dropout} or ensembles~\cite{DBLP:conf/nips/Lakshminarayanan17} are usually used to approximate Bayesian inference. The autoencoder (AE)~\cite{akcay2018ganomaly,baur2018deep} and RBM~\cite{creusot2015real} are the typical generative methods assuming that the reconstruction error of OOD images is larger than in-distribution images.

Recently, another kind of generative model-based on GAN resynthesis was proved to achieve state-of-the-art performance based on its reliable high resolution pixel-to-pixel translation results. SynthCP~\cite{xia2020synthesize} and DUIR~\cite{Lis2019} are two methods based on GAN resynthesis. Unfortunately, they are far away from lightweight as two or three neural networks have to be used in sequence for OOD detection. Compared to them, we demonstrate that the DMLNet based on contrastive clustering has better anomaly segmentation performance, while only needing to inference for one time.

\subsection{Deep metric learning networks}
DMLNets have been used in many kinds of applications, including video understanding~\cite{lee2018collaborative} and person re-identification~\cite{yi2014deep}. The DMLNet translates such problems to calculate embedded feature similarity in metric space using Euclidean, Mahalanobis, or Matusita distances~\cite{matusita1955decision}. Convolutional prototype networks and DMLNets are usually used together to tackle specific problems, such as detecting image-level OOD samples~\cite{yang2018robust, yang2020convolutional,chen2020learning} and few-shot learning for semantic segmentation~\cite{snell2017prototypical, oreshkin2018tadam, wang2018low}. We also follow this combination to build the first DMLNet for open world semantic segmentation.

\subsection{Open world classification and detection}
Open world classification was first proposed by~\cite{bendale2015towards}. This work presented the Nearest Non-Outlier (NNO) algorithm which is efficient in incrementally adding object categories, detecting outliers, and managing open space risk. Recently Joseph et al.~\cite{joseph2021open} proposed an open world object detection system based on contrastive clustering, an unknown-aware proposal network, and energy-based unknown identification criteria. The pipeline of our open world semantic segmentation system is similar to theirs except for two important differences that make our task more challenging: (1) In their open-set detection module, they rely on the fact that the Region Proposal Network (RPN) is class agnostic so that potential OOD objects which are not labeled can also be detected. In this way, the information of the OOD samples is valid for training. However, we focus on semantic segmentation where every pixel used in training is assigned an in-distribution label, so no OOD samples can be added into the training. (2) In the incremental learning module, they use all labeled data of the novel class, while we focus on the few-shot condition which is naturally more difficult. Little research focuses on the incremental few-shot learning, which includes incremental few-shot learning for classification~\cite{gidaris2018dynamic}, object detection~\cite{perez2020incremental} and semantic segmentation~\cite{cermelli2020few}.

\begin{figure*}[t]
\begin{center}
  \includegraphics[width=0.8\linewidth]{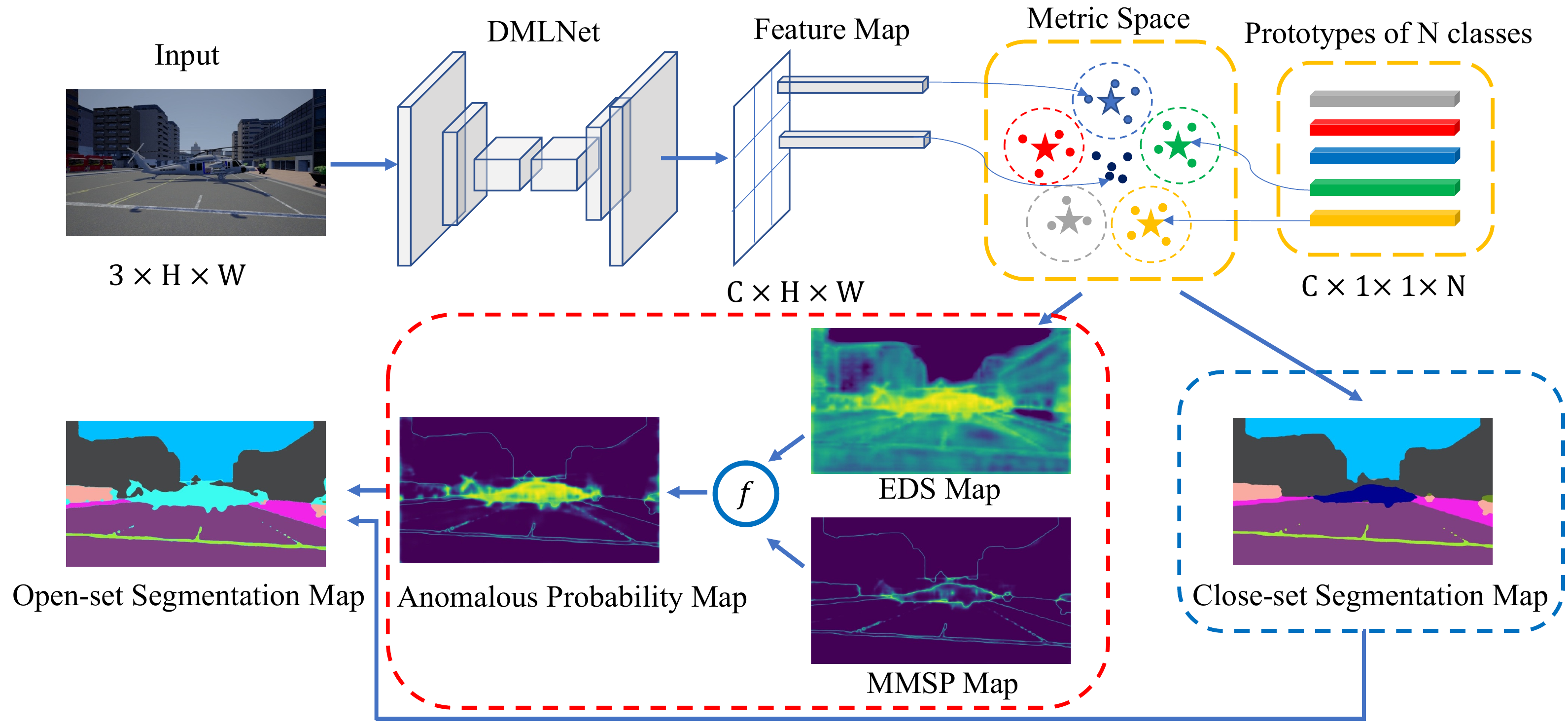}
\end{center}
\vspace{-0.2cm}
  \caption{{\bf Open-set semantic segmentation module.} The close-set segmentation submodule is enclosed within the blue dashed box, and the anomaly segmentation submodule is enclosed within the red dashed box. The open-set segmentation map is the combination of results generated by these two submodules. Both the in-distribution classes and OOD classes are predicted in the open-set segmentation map. Kindly refer to Section~\ref{sec:open-set} for the definition of the EDS map and MMSP map.}
\label{fig:second}
\vspace{-0.4cm}
\end{figure*}

\section{Open world semantic segmentation}
In this section, we give the working pipeline of the open world semantic segmentation system. This system is composed of an open-set semantic segmentation module and an incremental few-shot learning module. Suppose $\mathcal{C}_{in}=\left \{ C_{in,1},C_{in,2},...,C_{in,N} \right \}$ are $N$ in-distribution classes, which are all annotated in training datasets, and $\mathcal{C}_{out}=\left \{C_{out,1},C_{out,2},...,C_{out,M} \right \}$ are $M$ OOD classes not encountered in the training datasets.

The \textbf{open-set semantic segmentation module} is further divided into two submodules: a close-set semantic segmentation submodule and an anomaly segmentation submodule. We assume that $\mathbf{\hat{Y}}^{close}$ is the output map of the close-set semantic segmentation submodule, so the class of each pixel $\mathbf{\hat{Y}}^{close}_{i,j} \in \mathcal{C}_{in}$. The function of the anomaly segmentation submodule is to identify OOD pixels, and its output is called the anomalous probability map: $\mathbf{\hat P}\in [0,1]^{H \times W}$, where $H$ and $W$ denote the height and width of the input image. Based on $\mathbf{\hat{Y}}^{close}$ and $\mathbf{\hat P}$, the open-set semantic segmentation map $\mathbf{\hat{Y}}^{open}$ is given as:
\vspace{-0.2cm}
\begin{equation}
        \mathbf{\hat{Y}}^{open}_{i,j}=\left\{
        \begin{array}{cl}
        C_{anomaly} & {\mathbf{\hat P}_{i,j} > \lambda_{out}} \\
        \mathbf{\hat{Y}}^{close}_{i,j} & {\mathbf{\hat P}_{i,j} \leqslant \lambda_{out}}
    \end{array} \right.
    \label{con:open}
    \vspace{-0.2cm}
    \end{equation}
where $C_{anomaly}$ denotes the OOD class, and $\lambda_{out}$ is the threshold to determine OOD pixels. Therefore, the open-set semantic segmentation module is supposed to identify OOD pixels as well as assign correct in-distribution labels. Then $\mathbf{\hat{Y}}^{open}$ can be forwarded to labelers who can identify $C_{anomaly}$ from $\mathcal{C}_{out}$ and give corresponding annotations of the novel class.

The \textbf{incremental few-shot learning module} is used to update the knowledge base of the close-set segmentation submodule from $\mathcal{C}_{in}$ to $\mathcal{C}_{in+M}$ one by one when new labels are available, where $\mathcal{C}_{in+t}=\mathcal{C}_{in}\cup \left \{C_{out,1},C_{out,2},...,C_{out,t} \right \}, t\in \left \{1,2,...,M \right \}$. Fig.~\ref{fig:head} shows the circular working pipeline of the open world semantic segmentation system.

\begin{figure}[h]
\begin{center}
\setlength{\abovecaptionskip}{-2.cm}
   \includegraphics[width=1\linewidth]{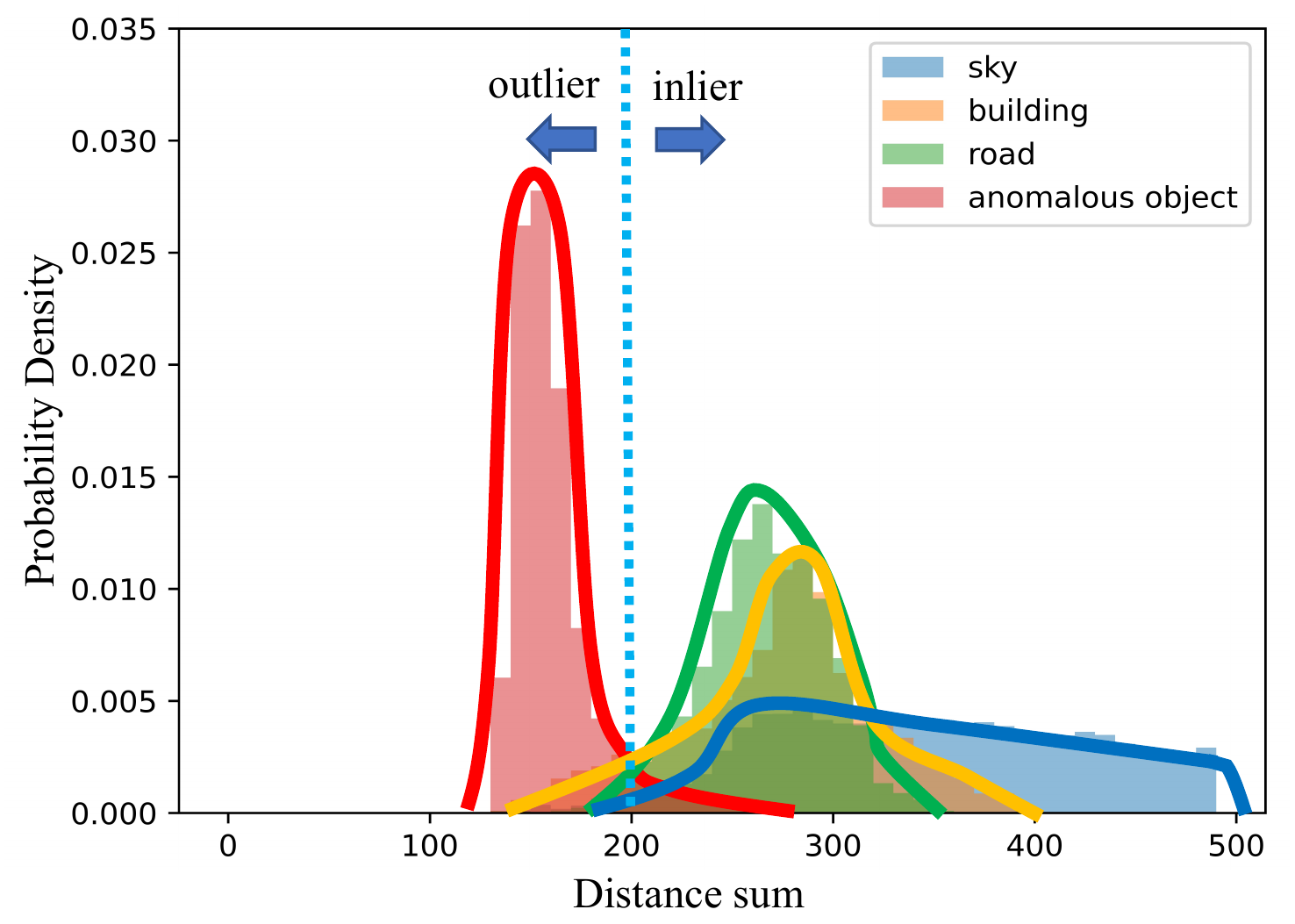}
   \caption{{\bf Distance sum distributions.} The statistic of this figure comes from the EDS map in Fig.~\ref{fig:second}. Only the classes whose number of pixels is over $4\%$ are demonstrated. This example demonstrates distinct peaks for inlier objects and outlier objects while anomalous classes have a smaller distance sum than known classes.}
\label{fig:third}
\end{center}
\vspace{-0.7cm}
\end{figure}

\section{Approach}

We adopt the DMLNet as our feature extractor and discuss the architecture and loss function in Section~\ref{sec:DMLNet}. The open-set segmentation module and the incremental few-shot learning module are illustrated in Section~\ref{sec:open-set} and~\ref{sec:incremental}.

\subsection{Deep Metric Learning Network}
\label{sec:DMLNet}
Classical CNN-based semantic segmentation networks can be disentangled into two parts: a feature extractor $f(\mathbf{X};\theta_f)$ for obtaining the embedding vector of each pixel and a classifier $g(f(\mathbf{X};\theta_f);\theta_g)$ for generating the decision boundary, where $\mathbf{X}$, $\theta_f$ and $\theta_g$ denote the input image, parameters of the feature extractor and classifier respectively. This learnable classifier is not suitable for OOD detection because it assigns all feature space to known classes and leaves no space for OOD classes. In contrast, the classifier is replaced by the Euclidean distance representation with all prototypes $\mathcal{M}_{in}=\left \{\mathbf{m}_t\in \mathbb{R}^{1 \times N}|t\in \left \{ 1,2,...,N \right \} \right \}$ in DMLNet, where $\mathbf{m}_t$ refers to the prototype of class $C_{in,t}$. The feature extractor $f(\mathbf{X};\theta_f)$ learns to map the input $\mathbf{X}$ to the feature vector which has the same length as the prototype in metric space. For the close-set segmentation task, the probability of one pixel $\mathbf{X}_{i,j}$ belonging to the class $C_{in,t}$ is formulated as:
\vspace{-0.1cm}
\begin{equation}
    p_t(\mathbf{X}_{i,j})=\frac{exp(-{\left \| f(\mathbf{X};\theta_f)_{i,j}-\mathbf{m}_t \right \|}^2)}{\sum_{t'=1}^{N}exp(-{\left \| f(\mathbf{X};\theta_f)_{i,j}-\mathbf{m}_{t'} \right \|}^2)} \label{con:probability}
\end{equation}

Based on this Euclidean distance-based probability, the \textbf{discriminative cross entropy (DCE)} loss function $\mathcal{L}_{\text{DCE}}(\mathbf{X}_{i,j},\mathbf{Y}_{i,j};\theta_f,\mathcal{M}_{in})$~\cite{yang2018robust} is defined as:
\vspace{-0.1cm}
\begin{equation}
\small
    \mathcal{L}_{\text{DCE}} =  \sum_{i,j} -log(\frac{exp(-{\left \| f(\mathbf{X};\theta_f)_{i,j}-\mathbf{m}_{\mathbf{Y}_{i,j}} \right \|}^2)}{\sum_{k=1}^{N}exp(-{\left \| f(\mathbf{X};\theta_f)_{i,j}-\mathbf{m}_{k} \right \|}^2)})
\end{equation}
where $\mathbf{Y}$ is the label of the input image $\mathbf{X}$. The numerator and denominator of $\mathcal{L}_{\text{DCE}}$ refer to the attractive force and repulsive force in Fig.~\ref{fig:contrastive} respectively. We formulate another loss function called the \textbf{variance loss (VL)} function $\mathcal{L}_{\text{VL}}(\mathbf{X}_{i,j},\mathbf{Y}_{i,j};\theta_f,\mathcal{M}_{in})$ which is defined as:
\vspace{-0.1cm}
\begin{equation}
    \mathcal{L}_{\text{VL}}=\sum_{i,j} {\left \| f(\mathbf{X};\theta_f)_{i,j}-\mathbf{m}_{\mathbf{Y}_{i,j}} \right \|}^2
    \vspace{-0.2cm}
\end{equation}

$\mathcal{L}_{\text{VL}}$ only has an attractive force effect but no repulsive force effect. With DCE and VL, the hybrid loss is defined as: $\mathcal{L}=\mathcal{L}_{\text{DCE}} + \lambda_{\text{VL}} \mathcal{L}_{\text{VL}}$, where $\lambda_{\text{VL}}$ is a weight parameter.

\subsection{Open-set semantic segmentation module}
\label{sec:open-set}
The open-set semantic segmentation module is composed of the close-set semantic segmentation submodule and anomaly segmentation submodule. The pipeline of the open-set semantic segmentation module is shown in Fig.~\ref{fig:second}. 

The \textbf{close-set semantic segmentation submodule} assigns in-distribution labels to all pixels of one image. As the probability of one pixel $\mathbf{X}_{i,j}$ belonging to class $C_{in,t}$ is formulated in Equation~\ref{con:probability}, the close-set segmentation map is:
\begin{equation}
    \mathbf{\hat Y}^{close}_{i,j}=\mathop{\arg\!\max_{t}} \ p_t(\mathbf{X}_{i,j})
    \label{con:close-set-map}
\end{equation}

The \textbf{anomaly segmentation submodule} detects OOD pixels. We propose two unknown identification criteria to measure the anomalous probability including \textit{metric-based maximum softmax probability (MMSP)} and \textit{Euclidean distance sum (EDS)}. Following is the anomalous probability based on the MMSP:
\vspace{-0.2cm}
\begin{equation}
    \mathbf{\hat P}^{\text{MMSP}}_{i,j}=1-max \ p_t(\mathbf{X}_{i,j}), \ t \in \left \{ 1,2,...,N \right \}
    \label{con:mmsp}
    \vspace{-0.2cm}
\end{equation}

The EDS is proposed from the finding that the Euclidean distance sum with all prototypes is smaller if the feature is located at the center of the metric space where OOD pixels aggregate. The EDS is defined as:
\vspace{-0.2cm}
\begin{equation}
    S(\mathbf{X}_{i,j})=\sum_{t=1}^{N}{\left \| f(\mathbf{X};\theta_f)_{i,j}-m_t \right \|}^2
    \vspace{-0.2cm}
\end{equation}

The anomalous probability based on the EDS is calculated as:
\vspace{-0.2cm}
\begin{equation}
    \mathbf{\hat P}^{\text{EDS}}_{i,j}=1-\frac{S(\mathbf{X}_{i,j})}{max \ {S(\mathbf{X})}}
    \label{con:EDS}
\end{equation}
\vspace{-0.3cm}

The EDS is class-independent, so prototypes of all classes should be evenly distributed in metric space and not moved during training. Learnable prototypes will cause instability during training and make no contribution to better performance~\cite{miller2020class}. Therefore, we define the prototype in a one-hot vector form: only the $t^{th}$ element of $m_t$ is $T$, while others remain zero, where $t\in \left \{1,2,...,N \right \}$.


Considering that the EDS is a ratio relative to the maximum distance sum among all pixels, the high anomalous score area definitely exists in every image even though there are no OOD objects in an image. Additionally, the distance sum distribution of every in-distribution class is slightly different from each other, as shown in Fig.~\ref{fig:third}. Therefore, we combine the MMSP with the EDS to suppress those pixels with middle response which are actually in-distribution. The mixture function is: 
\begin{equation}
    \mathbf{\hat P}=\alpha \  \mathbf{\hat P}^{\text{EDS}}+(1-\alpha) \ \mathbf{\hat P}^{\text{MMSP}}
    \label{con:mix}
    \vspace{-0.2cm}
\end{equation} and $\alpha$ is determined by:
\vspace{-0.3cm}
\begin{equation}
    \alpha = \frac{1}{1+exp(-\beta \ (\mathbf{\hat P}^{\text{EDS}}-\gamma))}
    \label{con:mixture}
    \vspace{-0.3cm}
\end{equation}
where $\beta$ and $\gamma$ are hyperparameters to control the suppressing effect and threshold. 

After obtaining the anomalous probability map by Equation~\ref{con:mix} and the close-set segmentation map by Equation~\ref{con:close-set-map}, we apply Equation~\ref{con:open} to generate the final open-set segmentation map.

\begin{figure}[t]
\begin{center}
  \includegraphics[width=1\linewidth]{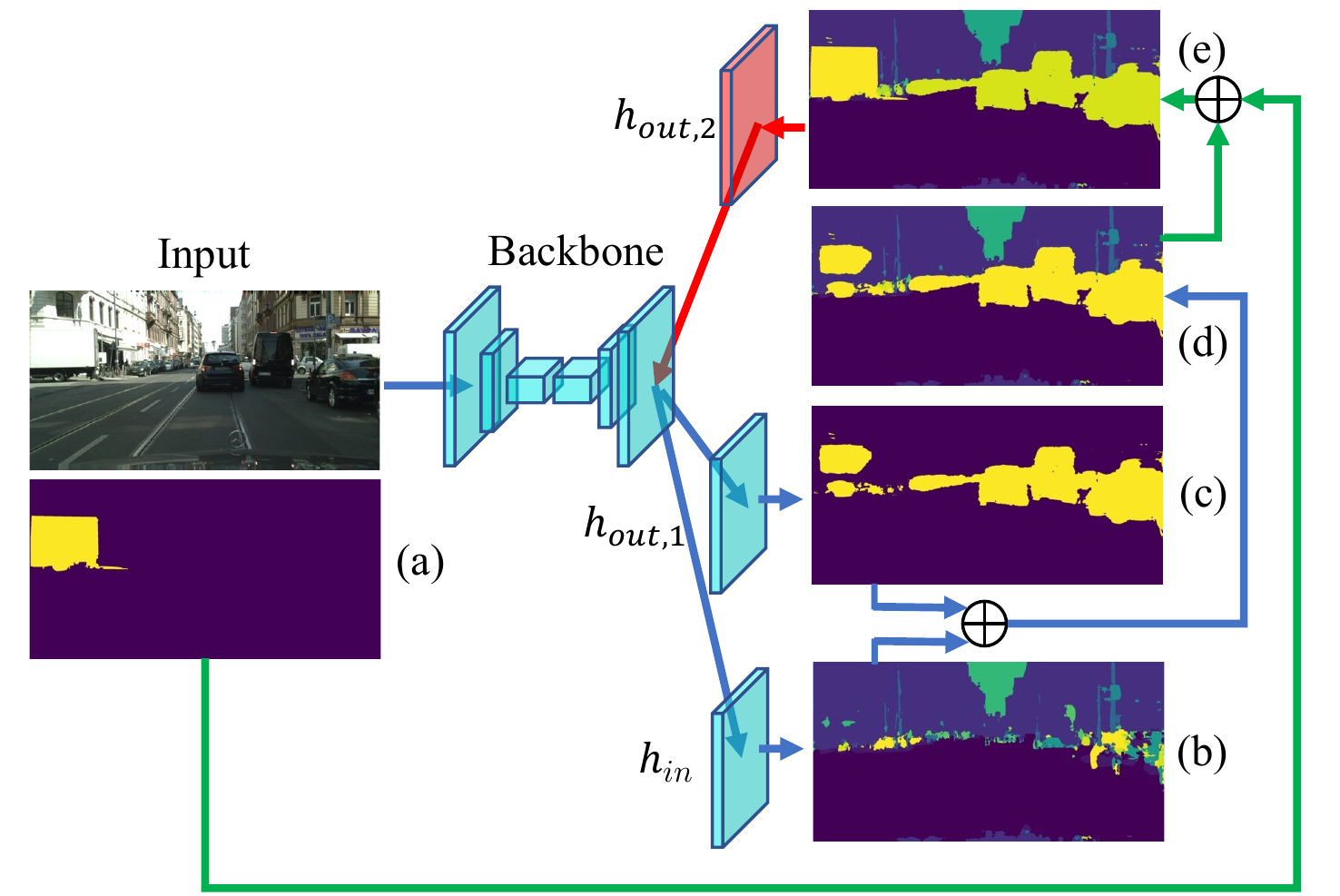}
  \caption{{\bf Pseudo Label Method.} This flowchart shows an example of how to train the new branch head $h_{out,2}$ which is in red. $h_{in}$ and $h_{out,1}$ provide the prediction maps (b) $\mathbf{M}_{in}$ and (c) $\mathbf{M}_{out,1}$ of the original in-distribution classes and the first learned OOD class respectively. Two of them are combined to generate (d) $\mathbf{M}_{in,1}$ which contains all known classes. Then the annotation map (a) $\mathbf{Y}_{out,2}$ of the novel class is merged to obtain the pseudo label map (e) $\mathbf{PL}_{in+1}$ for training. Different colors indicate different classes.}
  \label{fig:incremental}
  \end{center}
\vspace{-0.5cm}
\end{figure}

\subsection{Incremental few-shot learning module}
\label{sec:incremental}
The open-set segmentation results obtained from Section~\ref{sec:open-set} can be delivered to labelers who can give corresponding annotations for a certain new class. As the labeling process is extremely time-consuming, especially for segmentation tasks, we make following two assumptions: (1) Only the new class will be annotated while other pixels will be ignored. (2) Less than five images will be annotated.

We propose two methods for the DMLNet to implement incremental few-shot learning. The first method is called the \textbf{Pseudo Label Method (PLM)}, as shown in Fig.~\ref{fig:incremental}. In this method, DMLNet is divided into two parts: the final branch heads and the backbone. The trained branch heads will provide prediction maps of old classes, and labelers will provide the annotation of the new class. In this way, the whole image will be annotated and used to train the new branch head. The second method is called the \textbf{Novel Prototype Method (NPM)}. We find the prototype of the novel class directly using the original DMLNet. Based on the novel prototype and in-distribution prototypes, we develop several criteria to classify all classes.

\begin{figure}[t]
\centering
\fbox{\includegraphics[width=0.84\linewidth]{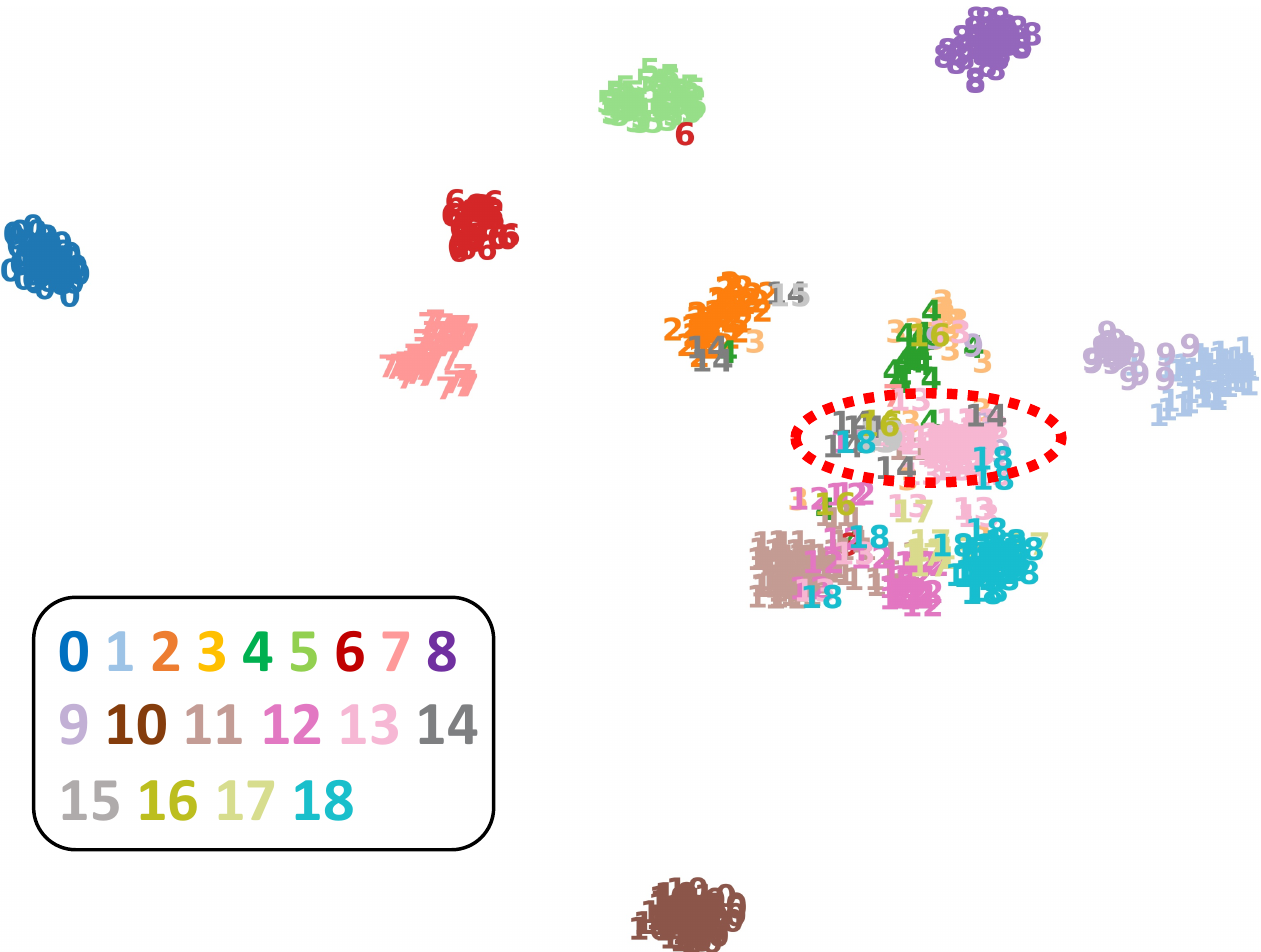}}
\vspace{0.15 cm}
\caption{\textbf{t-SNE visualization for learned metrics of open-set semantic segmentation module}. All learned metrics of 19 classes of the Cityscapes dataset are included. 13 (car), 14 (truck) and 15 (bus) are OOD classes and are enclosed within the red dotted circle. They are separated from known classes and they also aggregate closely in metric space.}
\label{fig:tsne}
\vspace{-0.3 cm}
\end{figure}

\noindent \textbf{Pseudo Label Method.} Suppose the DMLNet has $k+1$ trained final branch heads $\left \{ h_{in},h_{out,1},...,h_{out,k} \right \}$, 1 head $h_{out,k+1}$ to be trained and the backbone $B$. The initial head $h_{in}$ and backbone $B$ are trained using the large training dataset of $\mathcal{C}_{in}$ based on the method in Section~\ref{sec:DMLNet}. Therefore, $h_{in}$ provides the prediction map $\mathbf{M}_{in} \in \mathbb{R}^{H\times W}$ for in-distribution classes $\mathcal{C}_{in}$. Suppose $\left \{\mathbf{M}_{out,1},...,\mathbf{M}_{out,k} \right \}$ are binary maps generated by $\left \{h_{out,1},...,h_{out,k} \right \}$, and each of them is responsible for predicting only one particular OOD class, \eg, $\mathbf{M}_{out,t}$ is responsible for predicting $C_{out,t},t\in \left \{ 1,2,...,k\right \}$. $\mathbf{Y}_{out,k+1}$ is the binary annotation map of $C_{out,k+1}$ from the labelers. Algorithm~\ref{algorithm} is the procedure for generating the pseudo label $\mathbf{PL}_{in+k}$.

\vspace{-0.2cm}
\SetAlCapFnt{\small}
\SetAlFnt{\small}
\SetAlTitleFnt{\small}
\begin{algorithm}
\caption{Pseudo Label Generation Method}\label{algorithm}
\KwInput{$\mathbf{M}_{in}$, $\left \{\mathbf{M}_{out,1},...,\mathbf{M}_{out,k} \right \}, \mathbf{Y}_{out,k+1}$}
\KwOutput{\text{Pseudo Label Map} $\mathbf{PL}_{in+k}$}
$t \leftarrow 1$\;
\While{$t \leqslant k$}
{$\mathbf{M}_{in+t}\leftarrow \mathbf{M}_{in+t-1}$\;
$\mathbf{M}_{in+t}[\mathbf{M}_{out,t}=1] \leftarrow N+t$\;
$t \leftarrow t+1$}
$\mathbf{PL}_{in+k}\leftarrow \mathbf{M}_{in+k}$\;
$\mathbf{PL}_{in+k}[\mathbf{Y}_{out,k+1}=1] \leftarrow N+k+1$\;
\end{algorithm}
\vspace{-0.4cm}

$\mathbf{PL}_{in+k}$ is regarded as the ground truth and used to train the head $h_{in+k+1}$ based on the loss functions in Section~\ref{sec:DMLNet}. The dimension of the metric space will also increase one by one when we add the new head. When training the branch head $h_{in+k+1}$, the output feature of each pixel and all prototypes will have length $N+k+1$. In other words, the prototype set now becomes $\mathcal{M}_{in+k+1}=\left \{\mathbf{m}_t\in \mathbb{R}^{1 \times (N+k+1)}|t\in \left \{ 1,2,...,N+k+1 \right \} \right \}$. Note that when training the new branch head, the parameters of all the other heads and backbone will be frozen. Fig.~\ref{fig:incremental} gives the pipeline of training head $h_{out,2}$. The inference procedure is the same as how to generate a pseudo label map, as shown in Algorithm~\ref{algorithm} except that the last two steps are not needed. In this way, the old knowledge is preserved because the old branch still has a reliable ability to classify base classes.

\noindent \textbf{Novel Prototype Method.} This method is inspired by the finding that all feature vectors of certain OOD novel classes will also aggregate together in metric space. Fig.~\ref{fig:tsne} demonstrates that OOD features separate very well with in-distribution features while aggregate closely. Therefore, 
we determine the prototype of the novel class $C_{out,k}$ as the mean of all features belonging to this novel class:
\vspace{-0.2cm}
\begin{equation}
    \mathbf{m}_{N+k}=\frac{\sum_{q=1}^{Q}\sum_{i,j}^{}\mathbf{F}_{i,j}\cdot \mathbf{Y}_{i,j}}{\sum_{q=1}^{Q}\sum_{i,j}^{} \mathbf{Y}_{i,j}}
    \vspace{-0.2cm}
\end{equation}
where $Q$ is the number of annotated samples, and $\mathbf{Y}$ is the ground truth binary mask of novel class $C_{out,k}$, in which only those locations with novel class are 1, else 0. $\mathbf{F}=f(\mathbf{X};\theta_f)\in \mathbb{R}^{N\times H\times W}$ is the feature map generated by the DMLNet.

The new prototype set is no longer evenly distributed because original in-distribution prototypes are all one-hot vectors, but the newly added prototype of the novel class is not. So using Equation~\ref{con:probability} to classify novel classes is not suitable anymore. We develop these two criteria to decide whether one pixel belongs to $C_{out,k}$:
\vspace{-0.2cm}
\begin{equation}
\small
    \left\{
        \begin{array}{ll}
        {\left \| \mathbf{F}_{i,j}-\mathbf{m}_{N+k} \right \|}^2 < \lambda_{novel}\\
        {\left \| \mathbf{F}_{i,j}-\mathbf{m}_{N+k} \right \|}^2 < {\left \| \mathbf{F}_{i,j}-\mathbf{m}_{k'} \right \|}^2 & k'\neq N+k
    \end{array} \right.
    \vspace{-0.2cm}
\end{equation}
where $\lambda_{novel}$ is a threshold we can manually design. The intuition is that the novel pixel should be close enough with its corresponding prototype as well as closest to its corresponding prototype among all prototypes. This method also handles catastrophic forgetting very well because the model is not updated at all. There is no training procedure in this method. Simply calculating the novel prototype and evaluating the input image based on our criteria is the pipeline.

\section{Experiments}

Our experiments are divided into three parts. We first evaluate our open-set semantic segmentation approach in Section~\ref{sec:ex-open}. Then we demonstrate our incremental few-shot learning results in Section~\ref{sec:ex-incremental}. Based on the open-set semantic segmentation module and incremental few-show learning module, the whole open world semantic segmentation is realized in Section~\ref{sec:ex-open-world}.

\subsection{Open-set semantic segmentation}
\label{sec:ex-open}
\noindent \textbf{Datasets.} Three datasets including StreetHazards~\cite{hendrycks2019scaling}, Lost and Found~\cite{pinggera2016lost} and Road Anomaly~\cite{Lis2019} are used to testify the robustness and effectiveness of our DMLNet based open-set semantic segmentation method. Most anomalous objects of StreetHazards are large rare transportation machines such as helicopters, planes, and tractors. The Lost and Found dataset contains lots of small anomalous items like cargo, toys, and boxes. The Road Anomaly dataset no longer constrains the scenario within urban situations but also contains images of villages and mountains.

\noindent \textbf{Metrics.} Open-set semantic segmentation is the combination of close-set segmentation and anomaly segmentation as discussed in Section~\ref{sec:open-set}. For the close-set semantic segmentation task, we use mIoU to evaluate the performance. For the anomaly segmentation task, three metrics are used according to \cite{hendrycks2019scaling}, including area under ROC curve (AUROC), false-positive rate at $95\%$ recall (FPR95), and area under the precision-recall curve (AUPR).

\noindent \textbf{Implementation details.} For StreetHazards, we follow the same training procedure as ~\cite{hendrycks2019scaling} to train a PSPNet~\cite{Zhao2016} on the training set of StreetHazards. For Lost and Found and Road Anomaly, we follow~\cite{Lis2019} to use BDD100k~\cite{yu2018bdd100k} to train a PSPNet. Note that PSPNet is only used to extract features as we discussed in Section~\ref{sec:DMLNet}. $\lambda_{\text{VL}}$ of hybrid loss is 0.01. None zero element $T$ of all prototypes is 3. $\beta$ and $\gamma$ are 20 and 0.8 respectively in Equation~\ref{con:mixture}. 

\noindent \textbf{Baselines.} For StreetHazards, several baselines have been evaluated including MSP~\cite{hendrycks2016baseline}, Dropout~\cite{gal2016dropout}, AE~\cite{baur2018deep}, MaxLogit~\cite{hendrycks2019scaling} and SynthCP~\cite{xia2020synthesize}. For Lost and Found and Road Anomaly, baseline approaches contain MSP, MaxLogit, Ensemble~\cite{DBLP:conf/nips/Lakshminarayanan17}, RBM~\cite{creusot2015real} and DUIR~\cite{Lis2019}.

\noindent \textbf{Results.} The result of StreetHazards is demonstrated in Table~\ref{tab:1}. For Lost and Found and Road Anomaly, mIoU is invalid as they only provide OOD class labels but no specific in-distribution class labels. The results are in Table~\ref{tab:3}. Our experiments show that our DMLNet-based method achieves state-of-the-art performance in all three anomaly segmentation related metrics. Compared to recently proposed GAN-based approaches including DUIR and SynthCP, our method outperforms them in anomaly segmentation quality with a more lightweight structure, as they need two or three deep neural networks in the whole pipeline while we only need to inference for one time. The mIoU value of close-set segmentation in StreetHazards demonstrate that our method has no harm to close-set segmentation. Some qualitative results are shown in Fig.~\ref{fig:example}.

\noindent \textbf{Ablation studies.} We carefully conduct ablation experiments to study the effect of different loss functions (VL and DCE) and anomaly judgment criteria (EDS and MMSP), as shown in Table~\ref{tab:2}. The fact that DCE has better performance in mIoU than VL indicates the significance of the repulsive force. EDS outperforms MMSP a lot under all loss functions, meaning that the class-agnostic criterion is more suitable in the anomaly segmentation task than the class-dependent criterion.

\begin{table}[t]
\begin{center}
\small
\renewcommand\arraystretch{1}
\renewcommand\tabcolsep{8pt}
\begin{tabular}{lcccc}
\hline
\multicolumn{1}{l|}{Method}   & \multicolumn{1}{l}{AUPR$\uparrow$} & \multicolumn{1}{l}{AUROC$\uparrow$} & \multicolumn{1}{l}{FPR95$\downarrow$} & \multicolumn{1}{l}{mIoU$\uparrow$} \\ \hline
\multicolumn{1}{l|}{AE}       & 2.2                      & 66.1                      & 91.7                      & 53.2                      \\
\multicolumn{1}{l|}{MSP+CRF}      & 6.5                      & 88.1                      & 29.9                      & 53.2                     \\
\multicolumn{1}{l|}{MSP}      & 6.6                      & 87.7                      & 33.7                      &   53.2                   \\
\multicolumn{1}{l|}{Dropout}  & 7.5                      & 69.9                      & 79.4                      &   -                    \\
\multicolumn{1}{l|}{SynthCP}  & 9.3                      & 88.5                      & 28.4                      &    53.2                  \\
\multicolumn{1}{l|}{MaxLogit} & 10.6                     & 89.3                      & 26.5                      &      53.2                \\ \hline
\multicolumn{1}{l|}{DML} & \textbf{14.7}                     & \textbf{93.7}                      & \textbf{17.3}                      & \textbf{53.9}                     \\
\hline
\end{tabular}
\end{center}
\vspace{-0.2cm}
\caption{Open-set segmentation results on \textbf{StreetHazards}. All baselines except Dropout have the same close-set mIoU because they do not change the network structure nor the inference pipeline for the close-set segmentation submodule.}
\label{tab:1}
\vspace{-0.2cm}
\end{table}

\begin{table}[t]
\small
\renewcommand\arraystretch{1}
\renewcommand\tabcolsep{1pt}
\begin{center}
\begin{tabular}{l|cccccc}
\hline
Dataset & \multicolumn{3}{c|}{\textbf{Lost and Found}} & \multicolumn{3}{c}{\textbf{Road Anomaly}} \\ \hline
Method   & \multicolumn{1}{l}{AUPR$\uparrow$} & \multicolumn{1}{l}{AUROC$\uparrow$} & \multicolumn{1}{l|}{FPR95$\downarrow$} & \multicolumn{1}{l}{AUPR$\uparrow$} & \multicolumn{1}{l}{AUROC$\uparrow$} & \multicolumn{1}{l}{FPR95$\downarrow$} \\ \hline 
Ensemble       & -                      & 57                      & \multicolumn{1}{c|}{-}   & -                      & 67                      & -                         \\
RBM      & -                      & 86                      & \multicolumn{1}{c|}{-}        & -                      & 59                      & -                 \\
MSP      & 21                      & 83                      & \multicolumn{1}{c|}{31}    & 19                      & 70                      & 61                      \\
MaxLogit  & 37                      & 91                      & \multicolumn{1}{c|}{21}  & 32                      & 78                      & 49                         \\
DUIR  & -                      & 93                      & \multicolumn{1}{c|}{-}           & -                      & 83                      & -                                  \\ \hline
DML     & \textbf{45}            & \textbf{97}             & \multicolumn{1}{c|}{\textbf{10}}  & \textbf{37}            & \textbf{84}             & \textbf{37}      \\ \hline
\end{tabular}
\end{center}
\vspace{-0.2cm}
\caption{Anomaly segmentation results on \textbf{Lost and Found} and \textbf{Road Anomaly}.}
\label{tab:3}
\vspace{-0.2cm}
\end{table}

\begin{table}[t]
\begin{center}
\small
\renewcommand\arraystretch{1}
\renewcommand\tabcolsep{2pt}
\begin{tabular}{cccccccc}
\hline
VL       & \multicolumn{1}{c|}{DCE} & EDS & \multicolumn{1}{c|}{MMSP} & \multicolumn{1}{l}{AUPR$\uparrow$} & \multicolumn{1}{l}{AUROC$\uparrow$} & \multicolumn{1}{l}{FPR95$\downarrow$} & \multicolumn{1}{l}{mIoU$\uparrow$}  \\ \hline
\checkmark        & \multicolumn{1}{c|}{}    & \checkmark   & \multicolumn{1}{c|}{}     & \textbf{15.1} &91.5  &25.1  &\multirow{3}{*}{49.0}  \\
\checkmark        & \multicolumn{1}{c|}{}    &   & \multicolumn{1}{c|}{\checkmark}     & 6.8 &88.6  &27.7  &  \\
\checkmark        & \multicolumn{1}{c|}{}    & \checkmark & \multicolumn{1}{c|}{\checkmark}     & 13.7 &91.0  &25.5  &  \\ \hline
        & \multicolumn{1}{c|}{\checkmark}   & \checkmark    & \multicolumn{1}{c|}{}     & 13.3 &89.7  &27.8  &\multirow{3}{*}{52.6} \\ 
       & \multicolumn{1}{c|}{\checkmark}    &   & \multicolumn{1}{c|}{\checkmark }     & 8.2 &91.1  &21.3  &  \\
        & \multicolumn{1}{c|}{\checkmark}    & \checkmark   & \multicolumn{1}{c|}{\checkmark}     & 13.3 &92.4  &20.5  &  \\
         \hline
\checkmark        & \multicolumn{1}{c|}{\checkmark}   & \checkmark    & \multicolumn{1}{c|}{}     & 14.7 &\textbf{93.7}  &\textbf{17.3}  &\textbf{\multirow{3}{*}{53.9}} \\ 
\checkmark        & \multicolumn{1}{c|}{\checkmark}    &   & \multicolumn{1}{c|}{\checkmark }     & 8.8 &91.7  &19.8  &  \\
\checkmark        & \multicolumn{1}{c|}{\checkmark}    & \checkmark   & \multicolumn{1}{c|}{\checkmark}     & 14.1 &93.5  &18.0  &  \\
         \hline
\end{tabular}
\end{center}
\vspace{-0.2cm}
\caption{Ablation experiment results on \textbf{StreetHazards}. VL and DCE are loss functions when training the DMLNet. EDS and MMSP are unknown identification criteria for anomaly segmentation. Kindly refer to Section~\ref{sec:DMLNet} and Section~\ref{sec:open-set} for details.}
\label{tab:2}
\vspace{-0.2cm}
\end{table}

\subsection{Incremental few-shot learning}
\label{sec:ex-incremental}
Two incremental few-shot learning methods PLM and NPM are tested on the Cityscapes dataset. Car, truck, and bus are 3 OOD classes not involved in initial training while the other 16 classes are regarded as in-distribution classes. For PLM, the performance of the latest head PLM$\boldsymbol{\mathrm{_{latest}}}$ and all heads PLM$\boldsymbol{\mathrm{_{all}}}$ are both evaluated. $\lambda_{novel}$ is 0.15 to achieve the best performance of NPM. We show the incremental few-shot learning result Tab.~\ref{tab:4} of two settings: only adding car and adding car, truck, and bus one by one. The metric mIoU$\boldsymbol{\mathrm{_{harm}}}$ is the comprehensive index~\cite{Xian_2019_CVPR} that balances mIoU$\boldsymbol{\mathrm{_{old}}}$ and mIoU$\boldsymbol{\mathrm{_{novel}}}$. 

We see that PLM$\boldsymbol{\mathrm{_{all}}}$ performs better than PLM$\boldsymbol{\mathrm{_{latest}}}$ because the latest head only fits classes encountered during incremental few-shot learning, while older heads keep the information of prior knowledge. Even for classes involved in few new training samples, the latest head has worse results than all heads as the latest head cannot obtain robust features from only few samples. In most cases, PLM has the better performance in novel classes but the worse result in old classes than NPM, because the new added heads in PLM that are responsible for predicting novel classes are more likely to classify a pixel as a novel class so that more false-negative detections will be generated for old classes.

\begin{figure}[t]
\begin{center}
  \includegraphics[width=1\linewidth]{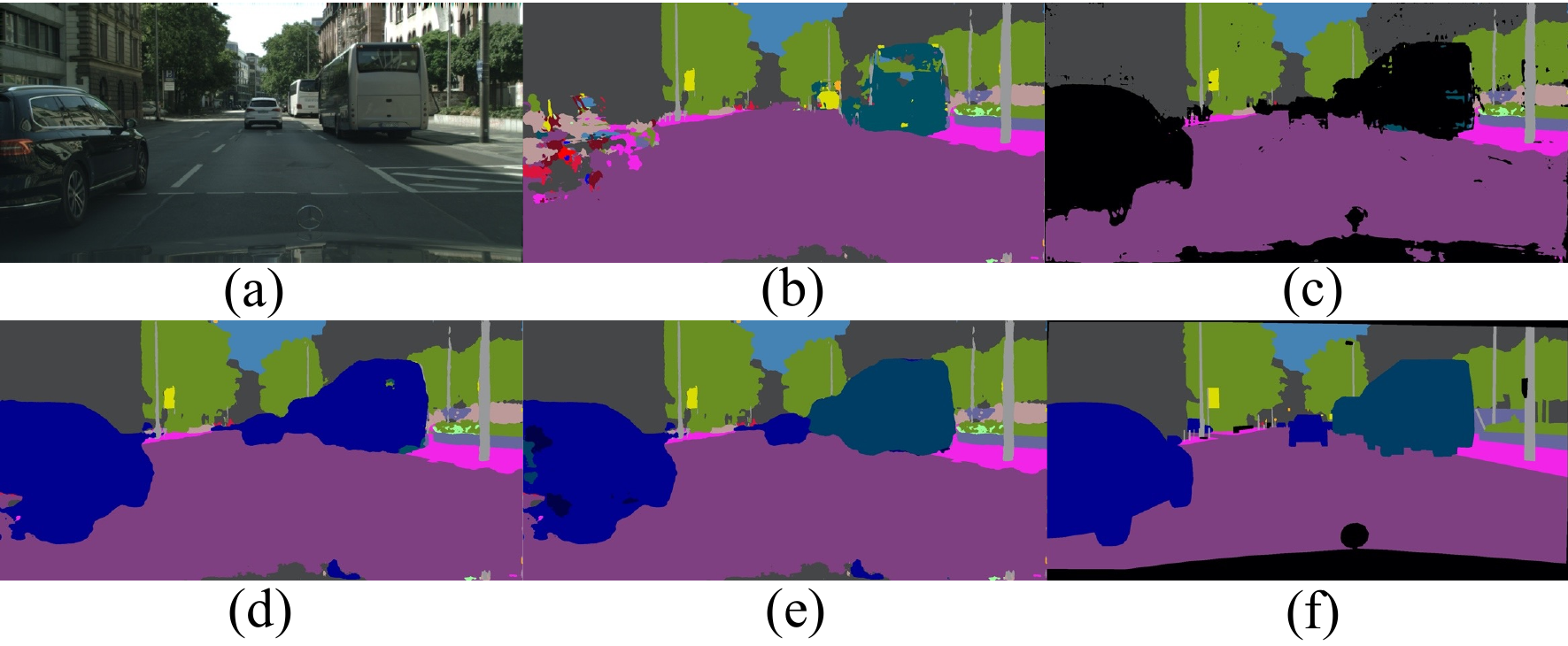}
\end{center}
\vspace{-0.3cm}
  \caption{Visualization results of our \textbf{open world semantic segmentation system}. (a) is input. (b) is close-set prediction. (c) is open-set prediction. (d) is prediction after 16+1 incremental few-shot learning. (e) is prediction after 16+3 incremental few-shot learning. (f) is ground truth.}
\label{fig:system}
\vspace{-0.5cm}
\end{figure}

\subsection{Open world semantic segmentation}
\label{sec:ex-open-world}
We realize an open world semantic segmentation system following these steps: \textbf{(1)} Train DMLNet for 16 classes of the Cityscapes dataset based on Section~\ref{sec:DMLNet}. Three classes including car, truck, and bus are excluded during training which is the same as Section~\ref{sec:ex-incremental}. \textbf{(2)} Obtain the close-set segmentation result based on Equation~\ref{con:close-set-map}. A visualization example is shown as (b) in Fig.~\ref{fig:system}. \textbf{(3)} Apply open-set semantic segmentation module based on Section~\ref{sec:open-set} to get open-set results like (c) in Fig.~\ref{fig:system}. The black area denotes anomalous objects. \textbf{(4)} Conduct incremental few-shot learning as discussed in Section~\ref{sec:incremental} to integrate car, truck, and bus one by one into existing perception knowledge. Results are shown as (d) and (e) in Fig.~\ref{fig:system}.

Close-set segmentation will assign an in-distribution label to OOD objects as shown in (b) of Fig.~\ref{fig:system}, which may cause safety-related accidents in autonomous driving. Open-set segmentation gives \textit{unknown} label to OOD objects, which not only makes the perception system more robust, but also provides information for labelers to make annotations of the new class. Incremental few-shot learning can be applied based on these new annotations to enlarge the perception knowledge base. From (d) and (e) of Fig.~\ref{fig:system} we can see our DMLNet can gradually classify car and bus. This \textit{open-set segmentation - incremental few-shot learning} cycle can progressively improve the performance of open world semantic segmentation system.

\begin{table*}[t]
\footnotesize
\renewcommand\arraystretch{1}
\renewcommand\tabcolsep{1.8pt}
\begin{tabular}{llccccccccccccccccccccccc}
\midrule
\color[HTML]{009901} \textbf{16 + 1 setting}& Method & \rotatebox{90}{road} & \rotatebox{90}{sidewalk} & \rotatebox{90}{building} & \rotatebox{90}{wall} & \rotatebox{90}{fence} & \rotatebox{90}{pole} & \rotatebox{90}{traffic light} & \rotatebox{90}{traffic sign} & \rotatebox{90}{vegetation} & \rotatebox{90}{terrain} & \rotatebox{90}{sky}  & \rotatebox{90}{person} & \rotatebox{90}{rider} & \rotatebox{90}{train} & \rotatebox{90}{motorcycle} & \rotatebox{90}{bicycle} & \rotatebox{90}{car} & \rotatebox{90}{truck} & \rotatebox{90}{bus} & \rotatebox{90}{mIoU} & \rotatebox{90}{mIoU$\boldsymbol{\mathrm{_{novel}}}$} & \rotatebox{90}{mIoU$\boldsymbol{\mathrm{_{old}}}$} & \rotatebox{90}{mIoU$\boldsymbol{\mathrm{_{harm}}}$} \\ \midrule
\multirow{3}{*}{{Baseline}}&All 17                                       & 97.8 & 82.4     & 91.8     & 52.3 & 57.5  & 59.9 & 64.1          & 74.2         & 91.9       & 61.4    & 94.6 & 79.4   & 58.8  & 75.6  & 61.7      & 74.9    & 94.8    & - & -                 & 74.9                      & -            & -          & -          \\
&First 16                                      & 98.0 & 82.1     & 91.4     & 43.6 & 56.4  & 58.9 & 61.4          & 72.6         & 91.6       & 60.5    & 94.4 & 79.1   & 57.6  & 67.9  & 61.1      & 75.1    & -       & - & -                 & 72.0                      & -            & -          & -          \\
&FT                                           & 0.0  & 0.0      & 0.0      & 0.0  & 0.0   & 0.0  & 0.0           & 0.0          & 0.0        & 0.0     & 0.0  & 0.0    & 0.0   & 0.0   & 0.0       & 0.0     & {\cellcolor[HTML]{DAE8FC}}6.6  & - & -                    & 0.4                       & 6.6          & 0.0        & 0.0        \\ \midrule
\multirow{3}{*}{5 shot}& PLM$\boldsymbol{\mathrm{_{latest}}}$                                  & 87.6 & 10.1     & 80.4     & {\cellcolor[HTML]{FCDCDA}0.7}   & 13.8  & 23.9 & {\cellcolor[HTML]{FCDCDA}}0           & 39.3         & 85.4       & {\cellcolor[HTML]{FCDCDA}}0.0     & 85.4 & 21.3   & 3.8   & {\cellcolor[HTML]{FCDCDA}2.1}   & {\cellcolor[HTML]{FCDCDA}0.0}       & 1.3     & {\cellcolor[HTML]{DAE8FC}75.7} & - & -                    & 31.2                      & \textbf{75.7}         & 28.4       & 41.3       \\
& PLM$\boldsymbol{\mathrm{_{all}}}$                                     & 97.1 & 79.3     & 89.2     & {\cellcolor[HTML]{FCDCDA}41.9} & 55.3  & 57.5 & \cellcolor[HTML]{FCDCDA}60.8          & 71.0         & 91.1       & \cellcolor[HTML]{FCDCDA}59.4    & 93.9 & 73.3   & 49.2  & \cellcolor[HTML]{FCDCDA}34.2  & \cellcolor[HTML]{FCDCDA}14.3       & 51.8    & \cellcolor[HTML]{DAE8FC}75.7   & - & -                  & 64.4                      & \textbf{75.7}         & 63.7       & \textbf{69.2}       \\
& NPM                                          & 96.2 & 79.3     & 89.2     & \cellcolor[HTML]{FCDCDA}41.6 & 52.0  & 56.3 & \cellcolor[HTML]{FCDCDA}61.1          & 69.4         & 90.4       & \cellcolor[HTML]{FCDCDA}58.8    & 94.1 & 74.4   & 55.3  & \cellcolor[HTML]{FCDCDA}53.4  & \cellcolor[HTML]{FCDCDA}39.2      & 70.3    & {\cellcolor[HTML]{DAE8FC}}64.6  & - & -                   & \textbf{67.4}                      & 64.6         & \textbf{67.6}       & 66.1   \\   \midrule
\multirow{3}{*}{1 shot}&PLM$\boldsymbol{\mathrm{_{latest}}}$ & 80.3 & \cellcolor[HTML]{FCDCDA}0.3  & 72.6 & \cellcolor[HTML]{FCDCDA}0.0  & \cellcolor[HTML]{FCDCDA}0.0  & \cellcolor[HTML]{FCDCDA}0.0  & \cellcolor[HTML]{FCDCDA}0.0  & 17.5 & 68.8 & \cellcolor[HTML]{FCDCDA}0.0  & 61.1 & \cellcolor[HTML]{FCDCDA}2.2  & 5.9  & \cellcolor[HTML]{FCDCDA}0.0  & \cellcolor[HTML]{FCDCDA}0.0  & 0.0  & {\cellcolor[HTML]{DAE8FC}}64.5 & - & - & 22.0 & \textbf{64.5} & 19.3 & 29.7 \\
&PLM$\boldsymbol{\mathrm{_{all}}}$    & 96.8 & \cellcolor[HTML]{FCDCDA}77.1 & 89.6 & \cellcolor[HTML]{FCDCDA}41.4 & \cellcolor[HTML]{FCDCDA}48.7 & \cellcolor[HTML]{FCDCDA}53.2 & \cellcolor[HTML]{FCDCDA}60.3 & 64.5 & 90.3 & \cellcolor[HTML]{FCDCDA}55.6 & 94.3 & \cellcolor[HTML]{FCDCDA}59.1 & 43.6 & \cellcolor[HTML]{FCDCDA}39.5 & \cellcolor[HTML]{FCDCDA}12.0 & 35.7 & {\cellcolor[HTML]{DAE8FC}}64.5 & - & - & 60.4 & \textbf{64.5} & 60.1 & 62.2 \\
&NPM         & 95.9 & \cellcolor[HTML]{FCDCDA}79.2 & 88.8 & \cellcolor[HTML]{FCDCDA}41.3 & \cellcolor[HTML]{FCDCDA}50.5 & \cellcolor[HTML]{FCDCDA}56.0 & \cellcolor[HTML]{FCDCDA}61.0 & 69.1 & 90.2 & \cellcolor[HTML]{FCDCDA}58.6 & 94.1 & \cellcolor[HTML]{FCDCDA}73.6 & 55.1 & \cellcolor[HTML]{FCDCDA}49.7 & \cellcolor[HTML]{FCDCDA}37.4 & 69.6 & {\cellcolor[HTML]{DAE8FC}}60.1 & - & - & \textbf{66.5} & 60.1 & \textbf{66.9} & \textbf{63.3} \\ \midrule
\color[HTML]{009901} \textbf{16 + 3 setting} & \\ \midrule
\multirow{3}{*}{{Baseline}}&All 19      & 97.9 & 83.0     & 91.7     & 51.5 & 58.3  & 59.8 & 64.2          & 74.2         & 92.0       & 61.2    & 94.6 & 79.7   & 59.1  & 63.9  & 61.5      & 75.0    & 94.2 & 78.5  & 81.4 & 74.8 &-              &-            &-            \\
&First 16    & 98.0 & 82.1     & 91.4     & 43.6 & 56.4  & 58.9 & 61.4          & 72.6         & 91.6       & 60.5    & 94.4 & 79.1   & 57.6  & 67.9  & 61.1      & 75.1    & -     &-   &-      & 72.0 &-              &-            &-            \\
&FT          & 0.0  & 0.0      & 0.0      & 0.0  & 0.0   & 0.0  & 0.0           & 0.0          & 0.0        & 0.0     & 0.0  & 0.0    & 0.0   & 0.0   & 0.0       & 0.0     & \cellcolor[HTML]{DAE8FC}0.0  & \cellcolor[HTML]{DAE8FC}0.0   & \cellcolor[HTML]{DAE8FC}0.4  & 0.0  & 0.1          & 0.0        & 0.0        \\ \midrule
\multirow{3}{*}{{5 shot}}&PLM$\boldsymbol{\mathrm{_{latest}}}$ & 91.6 & 58.3     & 78.4     & 7.4  & \cellcolor[HTML]{FCDCDA}3.3   & 34.1 & 34.0          & 42.7         & 85.9       & 23.4    & 86.2 & 5.4    & \cellcolor[HTML]{FCDCDA}0.0   & \cellcolor[HTML]{FCDCDA}0.0   & \cellcolor[HTML]{FCDCDA}0.0       & 3.1     & \cellcolor[HTML]{DAE8FC}63.8 & \cellcolor[HTML]{DAE8FC}1.3   & \cellcolor[HTML]{DAE8FC}12.0 & 33.2 & 25.7         & 34.6       & 29.5       \\
&PLM$\boldsymbol{\mathrm{_{all}}}$    & 97.1 & 79.2     & 84.8     & 38.1 & \cellcolor[HTML]{FCDCDA}46.4  & 56.8 & 58.8          & 61.0         & 91.0       & 59.3    & 92.9 & 63.6   & \cellcolor[HTML]{FCDCDA}47.5  & \cellcolor[HTML]{FCDCDA}3.4   & \cellcolor[HTML]{FCDCDA}13.8       & 47.5    & \cellcolor[HTML]{DAE8FC}67.0 & \cellcolor[HTML]{DAE8FC}5.7   & \cellcolor[HTML]{DAE8FC}12.0 & 54.0 & \textbf{28.2}         & 58.8       & \textbf{38.1}       \\
&NPM         & 96.1 & 79.3     & 58.7     & 41.5 & \cellcolor[HTML]{FCDCDA}51.5  & 56.3 & 60.7          & 69.0         & 90.4       & 58.8    & 94.1 & 74.3   & \cellcolor[HTML]{FCDCDA}55.1  & \cellcolor[HTML]{FCDCDA}32.0  & \cellcolor[HTML]{FCDCDA}39.1      & 70.2    & \cellcolor[HTML]{DAE8FC}55.7 & \cellcolor[HTML]{DAE8FC}1.6   & \cellcolor[HTML]{DAE8FC}21.0 & \textbf{58.2} & 26.1         & \textbf{64.2}       & 37.1       \\ \midrule
\multirow{3}{*}{{1 shot}}&PLM$\boldsymbol{\mathrm{_{latest}}}$ & 87.6 & 34.6 & 52.4 & \cellcolor[HTML]{FCDCDA}0.0  & \cellcolor[HTML]{FCDCDA}0.0  & 25.7 & 2.3  & 25.7 & 69.8 & 16.6 & \cellcolor[HTML]{FCDCDA}0.0  & \cellcolor[HTML]{FCDCDA}0.0  & \cellcolor[HTML]{FCDCDA}0.0  & \cellcolor[HTML]{FCDCDA}0.0  & \cellcolor[HTML]{FCDCDA}0.0  & \cellcolor[HTML]{FCDCDA}0.0  & \cellcolor[HTML]{DAE8FC}41.0 & \cellcolor[HTML]{DAE8FC}0.7 & \cellcolor[HTML]{DAE8FC}9.5  & 19.3 & 17.1 & 19.7 & 18.3 \\
&PLM$\boldsymbol{\mathrm{_{all}}}$    & 96.8 & 75.2 & 49.0 & \cellcolor[HTML]{FCDCDA}33.1 & \cellcolor[HTML]{FCDCDA}31.4 & 48.0 & 33.2 & 44.6 & 89.7 & 55.3 & \cellcolor[HTML]{FCDCDA}23.0 & \cellcolor[HTML]{FCDCDA}42.1 & \cellcolor[HTML]{FCDCDA}32.8 & \cellcolor[HTML]{FCDCDA}5.3  & \cellcolor[HTML]{FCDCDA}8.0  & \cellcolor[HTML]{FCDCDA}27.7 & \cellcolor[HTML]{DAE8FC}30.4 & \cellcolor[HTML]{DAE8FC}0.7 & \cellcolor[HTML]{DAE8FC}9.5  & 38.7 & 13.5 & 43.4 & 20.6 \\
&NPM         & 95.8 & 79.2 & 44.6 & \cellcolor[HTML]{FCDCDA}41.2 & \cellcolor[HTML]{FCDCDA}50.2 & 56.0 & 60.5 & 67.5 & 90.1 & 58.6 & \cellcolor[HTML]{FCDCDA}94.0 & \cellcolor[HTML]{FCDCDA}73.5 & \cellcolor[HTML]{FCDCDA}54.9 & \cellcolor[HTML]{FCDCDA}24.9 & \cellcolor[HTML]{FCDCDA}37.2 & \cellcolor[HTML]{FCDCDA}69.6 & \cellcolor[HTML]{DAE8FC}54.5 & \cellcolor[HTML]{DAE8FC}1.1 & \cellcolor[HTML]{DAE8FC}22.0 & \textbf{56.6} & \textbf{25.9} & \textbf{62.3} & \textbf{36.5} \\ \midrule
\end{tabular}
\setlength{\abovecaptionskip}{0.cm}
\caption{Incremental few-shot learning results on Cityscapes for 16 + 1 (car) setting and 16 + 3 (car, truck, bus) setting. The novel classes are in \colorbox[HTML]{DAE8FC}{blue} and the old classes not involved in new training samples are in \colorbox[HTML]{FCDCDA}{red}. Finetune (FT) is the baseline with catastrophic forgetting. The upper-bound is to regard the novel class as the original in-distribution class and retrain the model. All results are reported in IoU.}
\label{tab:4}
\vspace{-0.3cm}
\end{table*}

\begin{figure*}[h]
\begin{center}
  \includegraphics[width=1\linewidth]{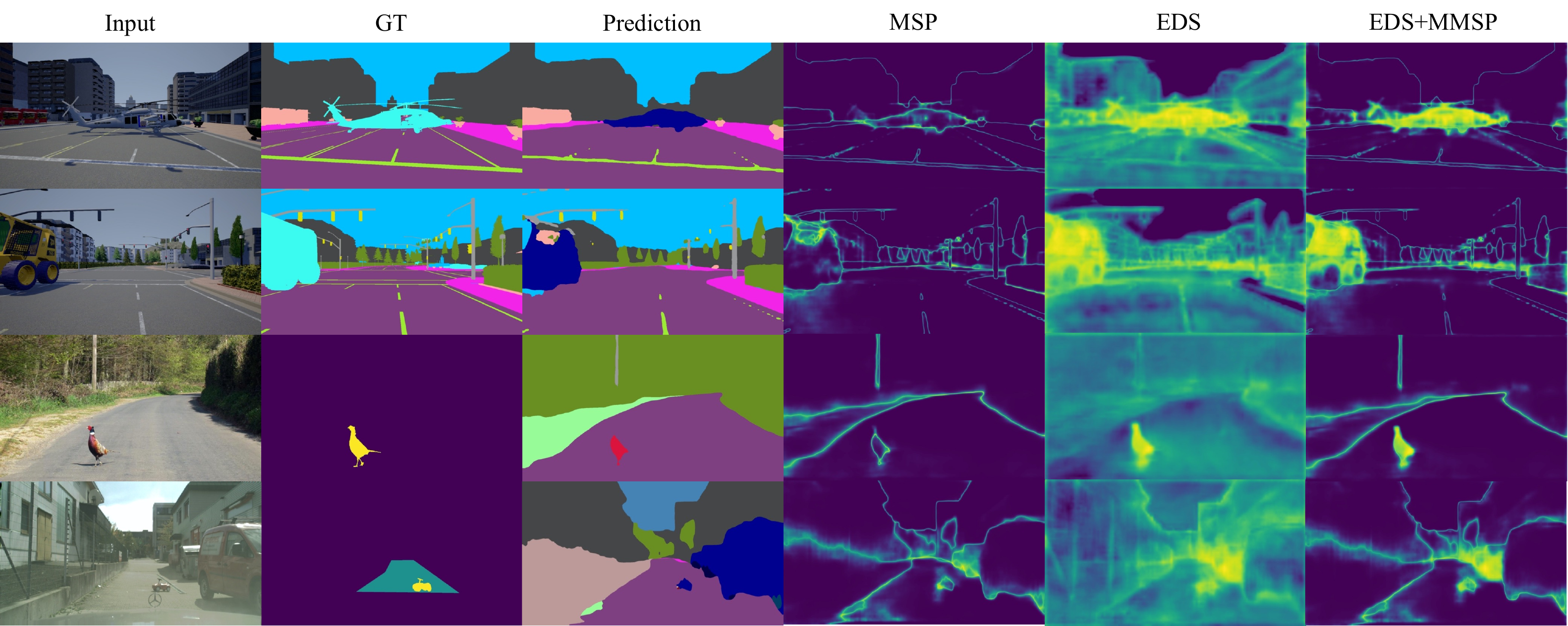}
\end{center}
\vspace{-0.3cm}
  \caption{Visualization results of \textbf{anomaly segmentation}. The first two rows are from \textbf{StreetHazards}, while the third and last rows are from \textbf{Road Anomaly} and \textbf{Lost and Found}. MSP is the baseline while EDS and EDS+MMSP are our methods.}
\label{fig:example}
\vspace{-0.5cm}
\end{figure*}

\section{Conclusion}

\vspace{-0.3cm}
We introduce an open world semantic segmentation system which incorporates two modules: an open-set segmentation module and an incremental few-shot learning module. Our proposed open-set segmentation module is based on the deep metric learning network, and it uses the Euclidean distance sum criterion to achieve state-of-the-art performance. Two incremental few-shot learning methods are proposed to broaden the perception knowledge of the network. Both modules of the open world semantic segmentation system can be further studied to improve the performance. We hope our work can draw more researchers to contribute to this practically valuable research direction.

\clearpage



{\small
\bibliographystyle{unsrt}
\bibliography{ms.bib}

\begin{thebibliography}{10}

\bibitem{chen2018encoder}
Liang-Chieh Chen, Yukun Zhu, George Papandreou, Florian Schroff, and Hartwig
  Adam.
\newblock Encoder-decoder with atrous separable convolution for semantic image
  segmentation.
\newblock In {\em Proceedings of the European conference on computer vision
  (ECCV)}, pages 801--818, 2018.

\bibitem{Zhao2016}
Hengshuang Zhao, Jianping Shi, Xiaojuan Qi, Xiaogang Wang, and Jiaya Jia.
\newblock Pyramid scene parsing network.
\newblock In {\em Proceedings of the IEEE/CVF Conference on Computer Vision and
  Pattern Recognition (CVPR)}, pages 2881--2890, 2017.

\bibitem{cordts2016cityscapes}
Marius Cordts, Mohamed Omran, Sebastian Ramos, Timo Rehfeld, Markus Enzweiler,
  Rodrigo Benenson, Uwe Franke, Stefan Roth, and Bernt Schiele.
\newblock The cityscapes dataset for semantic urban scene understanding.
\newblock In {\em Proceedings of the IEEE/CVF Conference on Computer Vision and
  Pattern Recognition (CVPR)}, pages 3213--3223, 2016.

\bibitem{huang2018apolloscape}
Xinyu Huang, Xinjing Cheng, Qichuan Geng, Binbin Cao, Dingfu Zhou, Peng Wang,
  Yuanqing Lin, and Ruigang Yang.
\newblock The apolloscape dataset for autonomous driving.
\newblock In {\em Proceedings of the IEEE Conference on Computer Vision and
  Pattern Recognition Workshops (CVPRW)}, pages 954--960, 2018.

\bibitem{Neuhold2017}
Gerhard Neuhold, Tobias Ollmann, Samuel Rota~Bulo, and Peter Kontschieder.
\newblock The mapillary vistas dataset for semantic understanding of street
  scenes.
\newblock In {\em Proceedings of the IEEE/CVF International Conference on
  Computer Vision (ICCV)}, pages 4990--4999, 2017.

\bibitem{janai2020computer}
Joel Janai, Fatma G{\"u}ney, Aseem Behl, Andreas Geiger, et~al.
\newblock Computer vision for autonomous vehicles: Problems, datasets and state
  of the art.
\newblock {\em Foundations and Trends{\textregistered} in Computer Graphics and
  Vision}, 12(1--3):1--308, 2020.

\bibitem{challen2019artificial}
Robert Challen, Joshua Denny, Martin Pitt, Luke Gompels, Tom Edwards, and
  Krasimira Tsaneva-Atanasova.
\newblock Artificial intelligence, bias and clinical safety.
\newblock {\em BMJ Quality \& Safety}, 28(3):231--237, 2019.

\bibitem{Bozhinoski2019}
Darko Bozhinoski, Davide~Di Ruscio, Ivano Malavolta, Patrizio Pelliccione, and
  Ivica Crnkovic.
\newblock Safety for mobile robotic system: A systematic mapping study from a
  software engineering perspective.
\newblock {\em Journal of Systems and Software}, 151:150--179, 2019.

\bibitem{hendrycks2016baseline}
Dan Hendrycks and Kevin Gimpel.
\newblock A baseline for detecting misclassified and out-of-distribution
  examples in neural networks.
\newblock In {\em International Conference on Learning Representations (ICLR)},
  2017.

\bibitem{gal2016dropout}
Yarin Gal and Zoubin Ghahramani.
\newblock Dropout as a bayesian approximation: Representing model uncertainty
  in deep learning.
\newblock In {\em international conference on machine learning (ICML)}, pages
  1050--1059, 2016.

\bibitem{hendrycks2019scaling}
Dan Hendrycks, Steven Basart, Mantas Mazeika, Mohammadreza Mostajabi, Jacob
  Steinhardt, and Dawn Song.
\newblock Scaling out-of-distribution detection for real-world settings.
\newblock {\em arXiv preprint arXiv:1911.11132}, 2019.

\bibitem{DBLP:conf/nips/Lakshminarayanan17}
Balaji Lakshminarayanan, Alexander Pritzel, and Charles Blundell.
\newblock Simple and scalable predictive uncertainty estimation using deep
  ensembles.
\newblock In {\em Advances in Neural Information Processing Systems (NeurIPS)},
  pages 6402--6413, 2017.

\bibitem{baur2018deep}
Christoph Baur, Benedikt Wiestler, Shadi Albarqouni, and Nassir Navab.
\newblock Deep autoencoding models for unsupervised anomaly segmentation in
  brain mr images.
\newblock In {\em International MICCAI Brainlesion Workshop}, pages 161--169,
  2018.

\bibitem{creusot2015real}
Clement Creusot and Asim Munawar.
\newblock Real-time small obstacle detection on highways using compressive rbm
  road reconstruction.
\newblock In {\em 2015 IEEE Intelligent Vehicles Symposium (IV)}, pages
  162--167, 2015.

\bibitem{bevandic2021dense}
Petra Bevandi{\'c}, Ivan Kre{\v{s}}o, Marin Or{\v{s}}i{\'c}, and Sini{\v{s}}a
  {\v{S}}egvi{\'c}.
\newblock Dense outlier detection and open-set recognition based on training
  with noisy negative images.
\newblock {\em arXiv preprint arXiv:2101.09193}, 2021.

\bibitem{Lis2019}
Krzysztof Lis, Krishna Nakka, Pascal Fua, and Mathieu Salzmann.
\newblock Detecting the unexpected via image resynthesis.
\newblock In {\em Proceedings of the IEEE/CVF International Conference on
  Computer Vision (ICCV)}, pages 2152--2161, 2019.

\bibitem{xia2020synthesize}
Yingda Xia, Yi~Zhang, Fengze Liu, Wei Shen, and Alan~L Yuille.
\newblock Synthesize then compare: Detecting failures and anomalies for
  semantic segmentation.
\newblock In {\em Proceedings of the European conference on computer vision
  (ECCV)}, pages 145--161. Springer, 2020.

\bibitem{mccloskey1989catastrophic}
Michael McCloskey and Neal~J Cohen.
\newblock Catastrophic interference in connectionist networks: The sequential
  learning problem.
\newblock In {\em Psychology of learning and motivation}, volume~24, pages
  109--165. Elsevier, 1989.

\bibitem{snell2017prototypical}
Jake Snell, Kevin Swersky, and Richard Zemel.
\newblock Prototypical networks for few-shot learning.
\newblock In {\em Advances in Neural Information Processing Systems (NeurIPS)},
  pages 4077--4087, 2017.

\bibitem{DBLP:conf/cvpr/RebuffiKSL17}
Sylvestre-Alvise Rebuffi, Alexander Kolesnikov, Georg Sperl, and Christoph~H
  Lampert.
\newblock icarl: Incremental classifier and representation learning.
\newblock In {\em Proceedings of the IEEE/CVF Conference on Computer Vision and
  Pattern Recognition (CVPR)}, pages 2001--2010, 2017.

\bibitem{mackay1995bayesian}
David~JC MacKay.
\newblock Bayesian neural networks and density networks.
\newblock {\em Nuclear Instruments and Methods in Physics Research Section A:
  Accelerators, Spectrometers, Detectors and Associated Equipment},
  354(1):73--80, 1995.

\bibitem{DBLP:conf/nips/KendallG17}
Alex Kendall and Yarin Gal.
\newblock What uncertainties do we need in bayesian deep learning for computer
  vision?
\newblock In {\em Advances in Neural Information Processing Systems (NeurIPS)},
  pages 5574--5584, 2017.

\bibitem{akcay2018ganomaly}
Samet Akcay, Amir Atapour-Abarghouei, and Toby~P Breckon.
\newblock Ganomaly: Semi-supervised anomaly detection via adversarial training.
\newblock In {\em Asian conference on computer vision}, pages 622--637, 2018.

\bibitem{lee2018collaborative}
Joonseok Lee, Sami Abu-El-Haija, Balakrishnan Varadarajan, and Apostol Natsev.
\newblock Collaborative deep metric learning for video understanding.
\newblock In {\em Proceedings of the 24th ACM SIGKDD International Conference
  on Knowledge Discovery \& Data Mining}, pages 481--490, 2018.

\bibitem{yi2014deep}
Dong Yi, Zhen Lei, Shengcai Liao, and Stan~Z Li.
\newblock Deep metric learning for person re-identification.
\newblock In {\em International Conference on Pattern Recognition}, pages
  34--39. IEEE, 2014.

\bibitem{matusita1955decision}
Kameo Matusita et~al.
\newblock Decision rules, based on the distance, for problems of fit, two
  samples, and estimation.
\newblock {\em The Annals of Mathematical Statistics}, 26(4):631--640, 1955.

\bibitem{yang2018robust}
Hong-Ming Yang, Xu-Yao Zhang, Fei Yin, and Cheng-Lin Liu.
\newblock Robust classification with convolutional prototype learning.
\newblock In {\em Proceedings of the IEEE/CVF Conference on Computer Vision and
  Pattern Recognition (CVPR)}, pages 3474--3482, 2018.

\bibitem{yang2020convolutional}
Hong-Ming Yang, Xu-Yao Zhang, Fei Yin, Qing Yang, and Cheng-Lin Liu.
\newblock Convolutional prototype network for open set recognition.
\newblock {\em IEEE Transactions on Pattern Analysis and Machine Intelligence
  (TPAMI)}, 2020.

\bibitem{chen2020learning}
Guangyao Chen, Limeng Qiao, Yemin Shi, Peixi Peng, Jia Li, Tiejun Huang,
  Shiliang Pu, and Yonghong Tian.
\newblock Learning open set network with discriminative reciprocal points.
\newblock In {\em Proceedings of the European conference on computer vision
  (ECCV)}, pages 507--522, 2020.

\bibitem{oreshkin2018tadam}
Boris~N. Oreshkin, Pau~Rodr{\'{\i}}guez L{\'{o}}pez, and Alexandre Lacoste.
\newblock {TADAM:} task dependent adaptive metric for improved few-shot
  learning.
\newblock In {\em Advances in Neural Information Processing Systems (NeurIPS)},
  pages 719--729, 2018.

\bibitem{wang2018low}
Yu-Xiong Wang, Ross Girshick, Martial Hebert, and Bharath Hariharan.
\newblock Low-shot learning from imaginary data.
\newblock In {\em Proceedings of the IEEE/CVF Conference on Computer Vision and
  Pattern Recognition (CVPR)}, pages 7278--7286, 2018.

\bibitem{bendale2015towards}
A.~{Bendale} and T.~{Boult}.
\newblock Towards open world recognition.
\newblock In {\em Proceedings of the IEEE/CVF Conference on Computer Vision and
  Pattern Recognition (CVPR)}, pages 1893--1902, 2015.

\bibitem{joseph2021open}
K~J Joseph, Salman Khan, Fahad~Shahbaz Khan, and Vineeth~N Balasubramanian.
\newblock Towards open world object detection.
\newblock In {\em Proceedings of the IEEE/CVF Conference on Computer Vision and
  Pattern Recognition (CVPR)}, 2021.

\bibitem{gidaris2018dynamic}
Spyros Gidaris and Nikos Komodakis.
\newblock Dynamic few-shot visual learning without forgetting.
\newblock In {\em Proceedings of the IEEE/CVF Conference on Computer Vision and
  Pattern Recognition (CVPR)}, pages 4367--4375, 2018.

\bibitem{perez2020incremental}
Juan-Manuel Perez-Rua, Xiatian Zhu, Timothy~M Hospedales, and Tao Xiang.
\newblock Incremental few-shot object detection.
\newblock In {\em Proceedings of the IEEE/CVF Conference on Computer Vision and
  Pattern Recognition (CVPR)}, pages 13846--13855, 2020.

\bibitem{cermelli2020few}
Fabio Cermelli, Massimiliano Mancini, Yongqin Xian, Zeynep Akata, and Barbara
  Caputo.
\newblock A few guidelines for incremental few-shot segmentation.
\newblock {\em arXiv preprint arXiv:2012.01415}, 2020.

\bibitem{miller2020class}
Dimity Miller, Niko Sunderhauf, Michael Milford, and Feras Dayoub.
\newblock Class anchor clustering: A loss for distance-based open set
  recognition.
\newblock In {\em Proceedings of the IEEE/CVF Winter Conference on Applications
  of Computer Vision (WACV)}, pages 3570--3578, 2021.

\bibitem{pinggera2016lost}
Peter Pinggera, Sebastian Ramos, Stefan Gehrig, Uwe Franke, Carsten Rother, and
  Rudolf Mester.
\newblock Lost and found: detecting small road hazards for self-driving
  vehicles.
\newblock In {\em IEEE/RSJ International Conference on Intelligent Robots and
  Systems (IROS)}, pages 1099--1106, 2016.

\bibitem{yu2018bdd100k}
Fisher Yu, Haofeng Chen, Xin Wang, Wenqi Xian, Yingying Chen, Fangchen Liu,
  Vashisht Madhavan, and Trevor Darrell.
\newblock {BDD100K:} {A} diverse driving dataset for heterogeneous multitask
  learning.
\newblock In {\em Proceedings of the IEEE/CVF Conference on Computer Vision and
  Pattern Recognition (CVPR)}, pages 2633--2642, 2020.

\bibitem{Xian_2019_CVPR}
Yongqin Xian, Subhabrata Choudhury, Yang He, Bernt Schiele, and Zeynep Akata.
\newblock Semantic projection network for zero- and few-label semantic
  segmentation.
\newblock In {\em Proceedings of the IEEE/CVF Conference on Computer Vision and
  Pattern Recognition (CVPR)}, June 2019.

\bibitem{he2016deep}
Kaiming He, Xiangyu Zhang, Shaoqing Ren, and Jian Sun.
\newblock Deep residual learning for image recognition.
\newblock In {\em Proceedings of the IEEE/CVF Conference on Computer Vision and
  Pattern Recognition (CVPR)}, pages 770--778, 2016.

\end{thebibliography}
}


\renewcommand\thesection{\Alph{section}}
\setcounter{section}{0}

\clearpage
\begin{center}
\large
    \textbf{Supplementary Material}
\end{center}

\section{Open-set semantic segmentation}

In this section, we provide these additional details for Section 5.1:
\vspace{-0.2cm}
\begin{itemize}
\setlength{\itemsep}{0pt}
\setlength{\parsep}{0pt}
\setlength{\parskip}{0pt}
    \item Details of each open-set semantic segmentation datasets.
    \item Details of open-set semantic segmentation implementation.
    \item Open-set semantic segmentation results under various $\beta$ and $\gamma$, which are hyperparameters of Equation 10. 
    \item Open-set semantic segmentation results under various $T$, which is the non-zero element of the prototypes.
\end{itemize}

\subsection{Datasets}
\textbf{StreetHazards} dataset~\cite{hendrycks2019scaling} contains 5125 images for training, 1031 images without anomalous objects for validation, and 1500 images for testing with anomalies. Twelve classes are involved during training, including sky, road, street lines, traffic signs, sidewalk, pedestrian, vehicle, building, wall, pole, fence, and vegetation. We include 250 unique anomaly models of diverse types in the test dataset, while most of them are large rare transportation machines.

\textbf{Lost and Found} dataset~\cite{pinggera2016lost} comprises 112 stereo video sequences with 2104 annotated frames. The whole dataset is only used for evaluating the anomaly segmentation performance and is not involved in training. 37 different obstacle types are contained and most of them are small items left on the street.

\textbf{Road Anomaly} dataset~\cite{Lis2019} is composed of 60 images for evaluating the anomalous objects. This dataset is no longer constrained in urban scenarios but contains images of villages and mountains. Animals, rocks, lost tires, trash cans are some anomalous examples in this dataset.

\subsection{Implementation}

For StreetHazards, we train a PSPNet decoder~\cite{Zhao2016} with a ResNet-101 encoder~\cite{he2016deep} for 20 epochs with batch size 8. We train both the encoder and decoder using SGD with the momentum of 0.9, the learning rate of $2\times10^{-2}$, and the learning rate decay of $10^{-4}$.

For Lost and Found and Road Anomaly, we use the training set from BDD100k~\cite{yu2018bdd100k} as these two datasets do not contain the training set themselves. The training procedure is as same as for StreetHazards because the number of training images in BDD100k is 4116, which is close to the number of training images in Streethazards.

\subsection{Varying $\beta$ and $\gamma$}

\begin{figure}[tb]
\begin{center}
  \includegraphics[width=1\linewidth]{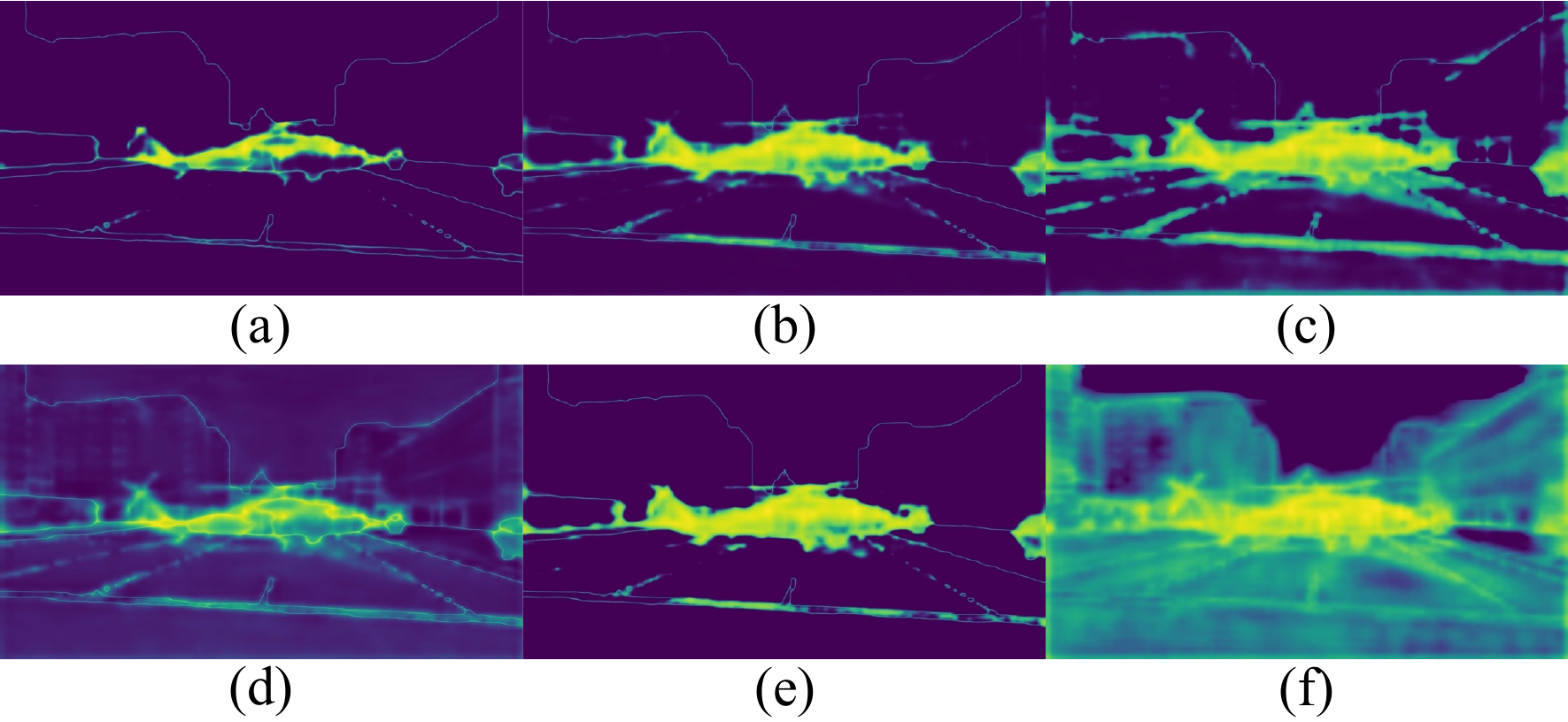}
\end{center}
  \caption{\textbf{Visualization results of different $\beta$ and $\gamma$}. (a) $\beta=50,\gamma=0.9$. (b) $\beta=50,\gamma=0.7$. (c) $\beta=50,\gamma=0.5$. (d) $\beta=5,\gamma=0.8$. (e) $\beta=20,\gamma=0.8$. (f) EDS only. From these visualization results we can see that MMSP can suppress the middle response of EDS.}
\label{fig:a}
\end{figure}

Equation 10 describes the way of using MMSP to suppress the middle response of EDS. Pixels whose EDS score is smaller than $\gamma$ are suppressed by MMSP, and $\beta$ controls the suppressing effect. The ablation experiment results are in Table~\ref{tab:beta_gamma}. Some qualitative results are shown in Fig.~\ref{fig:a}.

From Fig.~\ref{fig:a} we can see that OOD objects are more obvious using Equation 10, but the fact is the mixture of EDS and MMSP provides similar metrics to EDS alone according to Table~\ref{tab:beta_gamma}. This is because: (1) MMSP will not only suppress in-distribution pixels, but also OOD pixels. For example, in (a) of Fig.~\ref{fig:a}, some pixels of the helicopter are also suppressed.  (2) All three anomaly segmentation related metrics are threshold-independent and used to measure whether anomalous scores of OOD pixels and in-distribution pixels are distinguishable, not the absolute difference value. EDS is already able to differentiate OOD pixels and in-distribution as shown in the Fig. 4 of our manuscript. However, the mixture map of EDS and MMSP can give labelers a better view and tells them the location of OOD objects, so they can make annotations more easily and pass them to our next incremental few-shot learning module. 

\begin{table}[tb]
\begin{center}
\small
\renewcommand\arraystretch{1.1}
\renewcommand\tabcolsep{5pt}
\begin{tabular}{cc|ccc|c}
\hline
$\gamma$  & $\beta$  & AUPR$\uparrow$ & AUROC$\uparrow$ & FPR95$\downarrow$ & mIoU$\uparrow$ \\ \hline
$\times$                   &$\times$    & \textbf{14.7} & \textbf{93.7}  & 17.3  & \multirow{9}{*}{53.9}     \\ \cline{1-5}
0.9                  & \multirow{5}{*}{20} & 12.4 & 93.0  & 15.9  &      \\
0.8                  &    & 14.1 & 93.5  & 18.0  &      \\
0.7                  &    & 14.6 & 93.6  & 17.5  &      \\
0.6                  &    & \textbf{14.7} & \textbf{93.7}  & 17.3  &      \\
0.5                  &    &  \textbf{14.7}    & \textbf{93.7}      & \textbf{17.2}      &      \\ \cline{1-5}
\multirow{3}{*}{0.8} & 5  & 12.0     & 93.2      & 17.9      &      \\
                     & 20 & 14.1 & 93.5  & 18.0  &      \\
                     & 50 &  14.4    &  93.4     &  18.6     &      \\ \hline
\end{tabular}
\end{center}
\caption{\textbf{Ablation experiment results of $\beta$ and $\gamma$}. The unknown identification criterion of the first row is EDS without MMSP. In our experiments, mIoU values are same because the close-set segmentation submodule is not influenced by Equation 10. It is shown that $\beta$ and $\gamma$ do not have huge impact on the performance.}
\label{tab:beta_gamma}
\end{table}

\subsection{Varying the non-zero element $T$ of prototypes}

$T$ is the non-zero element of all prototypes as discussed in Section 4.2. It controls the positions of all prototypes in the metric space. Here we vary $T$, while the loss function is hybrid loss and the unknown identification criterion is EDS. The result is Table~\ref{tab:T}. We find that different $T$ has similar close-set mIoU indicating that $T$ has little influence on the close-set segmentation. The most related metric among all anomaly segmentation metrics is FPR95, meaning that the appropriate $T$ can reduce the false-positive detection.

\begin{table}[h]
\begin{center}
\small
\renewcommand\arraystretch{1}
\renewcommand\tabcolsep{5pt}
\begin{tabular}{lcccc}
\hline
\multicolumn{1}{l|}{$T$}   & \multicolumn{1}{l}{AUPR$\uparrow$} & \multicolumn{1}{l}{AUROC$\uparrow$} & \multicolumn{1}{l}{FPR95$\downarrow$} & \multicolumn{1}{l}{mIoU$\uparrow$} \\ \hline
\multicolumn{1}{l|}{1}       & 14.2                      & 88.1                      & 35.1                      & 53.6                      \\
\multicolumn{1}{l|}{2}      & 14.9                      & 92.2                      & 22.0                      & \textbf{54.1}                     \\
\multicolumn{1}{l|}{3}      & 14.7                      & 93.7                      & \textbf{17.3}                      &   53.9                   \\
\multicolumn{1}{l|}{4}  & \textbf{15.0}                      & \textbf{93.9}                      & 17.4                      &   53.9                   \\
\multicolumn{1}{l|}{5}  & 14.1                      & 93.4                      & 19.0                      &    53.8                  \\
\multicolumn{1}{l|}{6} & 13.7                     & 93.6                      & 21.4                      &      53.8                \\ 
\hline
\end{tabular}
\end{center}
\caption{\textbf{Ablation experiment results of $T$}. We find the DMLNet has nice anomaly segmentation performance when $T=3$ and $T=4$.}
\label{tab:T}
\end{table}

\section{Incremental few-shot learning}

In this section, we provide the following details for Section 5.2:

\vspace{-0.2cm}
\begin{itemize}
\setlength{\itemsep}{0pt}
\setlength{\parsep}{0pt}
\setlength{\parskip}{0pt}
    \item Details of the network architecture and training implementation.
    \item Incremental few-shot learning results of the novel prototype method (NPM) under various $\lambda_{novel}$ of the Equation 12.
    \item Incremental learning under the few-shot and non-few-shot condition.
    \item Incremental learning using the pseudo labels and ground truth labels.
\end{itemize}

\subsection{Implementation}

\begin{figure}[t]
\begin{center}
  \includegraphics[width=0.9\linewidth]{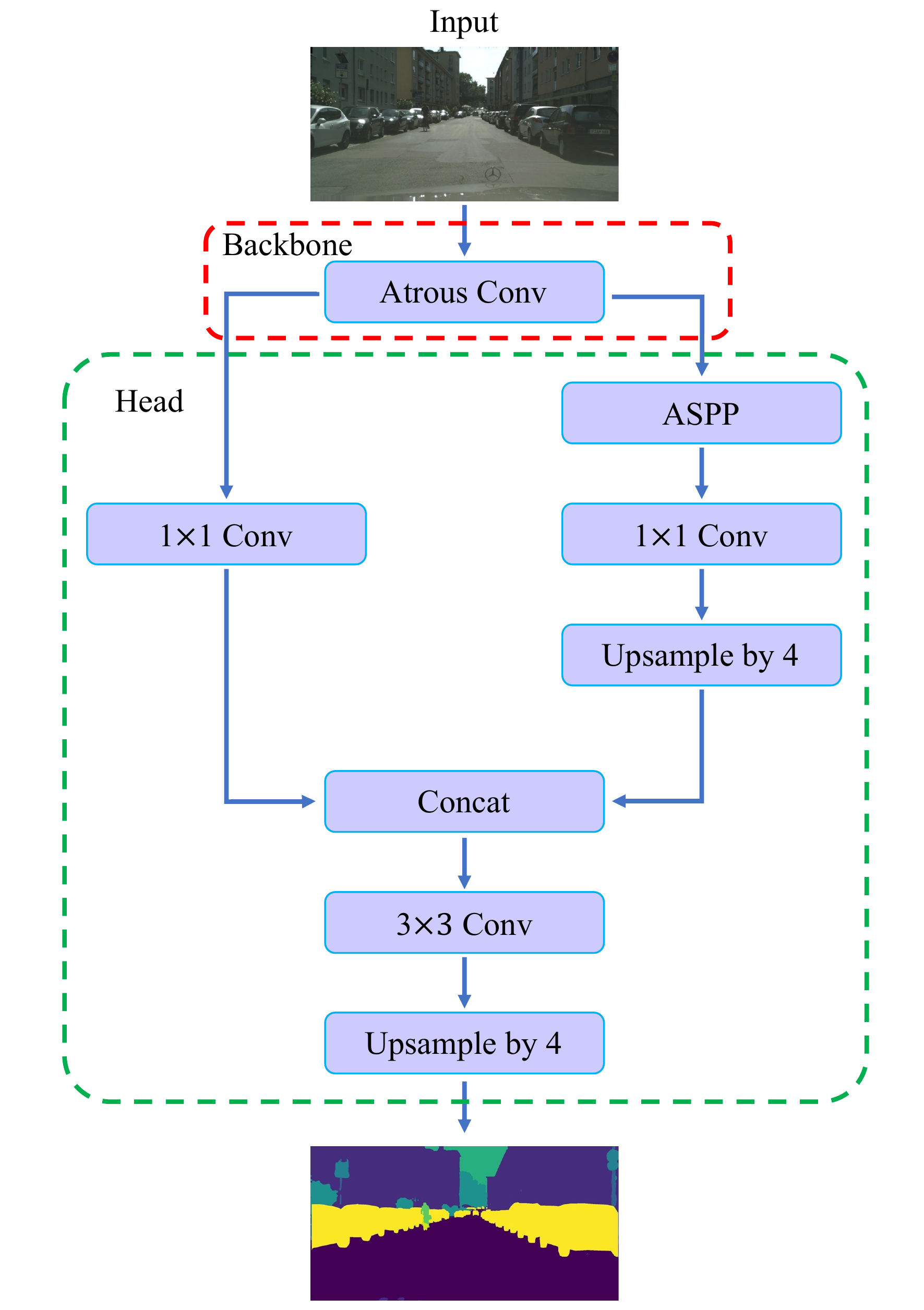}
\end{center}
\vspace{-0.3cm}
  \caption{\textbf{DMLNet architecture based on DeeplabV3+}. This is the network architecture we adopt for the incremental few-shot learning module. The backbone and head in Fig. 5 of our manuscript are demonstrated specifically in this Fig.~\ref{fig:b}.}
\label{fig:b}
\vspace{-0.5cm}
\end{figure}

The DMLNet we adopt for incremental few-shot learning is based on DeeplabV3+~\cite{chen2018encoder}, as shown in Fig.~\ref{fig:b}.

We train the base model on the Cityscapes dataset~\cite{cordts2016cityscapes} containing high quality pixel-level annotations of 5000 images (2975 and 500 for the training and validation respectively). The labels of 3 classes including car, truck and bus are set to be 255, so they are ignored during training. We train the encoder and decoder using SGD with the momentum of 0.9, the learning rate decay of $10^{-4}$, and the initial learning rate of 0.01 and 0.1 respectively for $3\times 10^{4}$ iterations. The batch size is 8 and the crop size is 762 due to the GPU memory limitation.

For the novel prototype method (NPM), we do not have to retrain the model for incremental few-shot learning as discussed in Section 4.3. For the pseudo label method (PLM), the architecture of the backbone and final branch head are demonstrated in Fig.~\ref{fig:b}. When we apply the PLM for each novel class, we fix the trained backbone and heads and decrease the initial learning rate to 0.01 and 0.001 for 5 shot and 1 shot respectively. Total iteration numbers are both 500 for 5 shot and 1 shot but the batch size is 5 and 1 respectively.

\subsection{Varying the $\lambda_{novel}$ of NPM}

The $\lambda_{novel}$ in Equation 12 controls the distance threshold for the novel class classification of NPM. We conduct the ablation experiments for various $\lambda_{novel}$ and the results are in Table~\ref{tab:lamda}. Large $\lambda_{novel}$ will cause more false-positive detection while small $\lambda_{novel}$ will cause more false-negative detection for the novel class. $\lambda_{novel}=1.5$ achieves the best performance according to Table~\ref{tab:lamda}.

\begin{table}[t]
\small
\renewcommand\arraystretch{1.1}
\renewcommand\tabcolsep{3pt}
\begin{center}
\begin{tabular}{lccccc}
\hline
\color[HTML]{009901} \textbf{16+1 setting}            & $\lambda_{novel}$ & mIoU & mIoU$\boldsymbol {\mathrm{_{novel}}}$ & mIoU$\boldsymbol{\mathrm{_{old}}}$  & mIoU$\boldsymbol{\mathrm{_{harm}}}$ \\ \hline
\multirow{3}{*}{5 shot} & 2     & \textbf{67.8} & 61.9  & \textbf{68.2} & 64.9 \\
                        & 1.5   & 67.4 & \textbf{64.6}  & 67.6 & \textbf{66.1} \\
                        & 1     & 63.4 & 41.8  & 64.7 & 50.8 \\ \hline
\multirow{3}{*}{1 shot} & 2     & \textbf{67.1} & 55.4  & \textbf{67.9} & 61.0 \\
                        & 1.5   & 66.5 & \textbf{60.1}  & 66.9 & \textbf{63.3} \\
                        & 1     & 62.5 & 38.0  & 64.1 & 47.7 \\ \hline
\color[HTML]{009901} \textbf{{16+3 setting}}                             \\ \hline
\multirow{3}{*}{5 shot} & 2     & 55.4 & 24.1  & 61.3 & 34.6 \\
                        & 1.5   & \textbf{58.2} & \textbf{26.1}  & \textbf{64.2} & \textbf{37.1} \\
                        & 1     & 56.6 & 20.2  & 63.4 & 30.7 \\ \hline
\multirow{3}{*}{1 shot} & 2     & 54.6 & 24.2  & 60.3 & 34.6 \\
                        & 1.5   & \textbf{56.6} & \textbf{25.9}  & 62.3 & \textbf{36.5} \\
                        & 1     & 55.5 & 18.9  & \textbf{62.4} & 29.0 \\ \hline
\end{tabular}
\end{center}
\caption{\textbf{Ablation experiment results of $\lambda_{novel}$}. We find that $\lambda_{novel}=1.5$ has the best performance among all settings.}
\label{tab:lamda}
\end{table}

\subsection{Few-shot and non-few-shot}

In the paper, we increase the knowledge base of the DMLNet through the incremental few-shot learning module. This is because: (1) Few-shot learning requires much fewer labels compared to non-few-shot learning, while making segmentation labels is extremely time-consuming. (2) The training process of the few-shot learning also consumes less time. (3) Incremental few-shot learning is not well studied so far, and we provide two methods including PLM and NPM as the baseline in this area. 

However, PLM and NPM perform worse than the upper bound which regards the novel class as one of the original in-distribution classes and retrain the model using the whole dataset. As the upper bound is under the non-few-shot condition, there are two possible reasons that make our incremental few-shot learning methods perform worse than the upper bound. The first one is that the training samples are insufficient, so the DMLNet cannot extract representative features. The second one is that our methods themselves constrain the DMLNet to obtain good performance. Therefore, we conduct experiments using more training samples to find out the reason.

From Table 4 of our manuscript, we notice that PLM have a better performance on the novel class than NPM under 5 shot condition. This is because the network architecture and the metric space of PLM will grow to fit the new classes. Therefore, PLM is more suitable for the non-few-shot condition. We conduct ablation experiments of PLM using a different number of training samples $Q$. The results are shown in Table~\ref{tab:incre}. We find the performance of PLM improves with more training samples, but the performance still not reaches the upper bound when using the whole training set of the Cityscapes dataset. Therefore, both the limited number of training samples and the PLM itself constrain the performance under few-shot condition.
\vspace{-0.1cm}

\begin{table}[h]
\small
\renewcommand\arraystretch{1.1}
\renewcommand\tabcolsep{3pt}
\begin{center}
\begin{tabular}{lccccc}
\hline
\color[HTML]{009901} \textbf{16+1 setting} & mIoU & mIoU$\boldsymbol {\mathrm{_{novel}}}$ & mIoU$\boldsymbol{\mathrm{_{old}}}$  & mIoU$\boldsymbol{\mathrm{_{harm}}}$ \\ \hline
{$Q=1$ }  & {60.4} & 64.5  & {60.1} & 62.2 \\
{$Q=5$ }  & 64.4 & {75.7}  & 63.7 & 69.2 \\
{$Q=100$ }& 70.7     & 85.5 & 69.8  & 76.9 \\
{$Q=1000$ }& 71.9     & 90.1 & 70.7  & 79.2 \\
{$Q=2975$ }& \textbf{72.2}     & \textbf{91.8} & \textbf{71.0}  & \textbf{80.1} \\ \hline
\textbf{All 17}& 74.9     & 94.8 & -  & - \\
\hline
\end{tabular}
\end{center}
\caption{\textbf{Ablation experiment results of $Q$}. \textbf{All 17} is the upper bound. The performance of PLM improves with more training samples.}
\label{tab:incre}
\vspace{-0.2cm}
\end{table}

\subsection{Pseudo label and ground truth label}

\begin{figure}[h]
\begin{center}
  \includegraphics[width=0.8\linewidth]{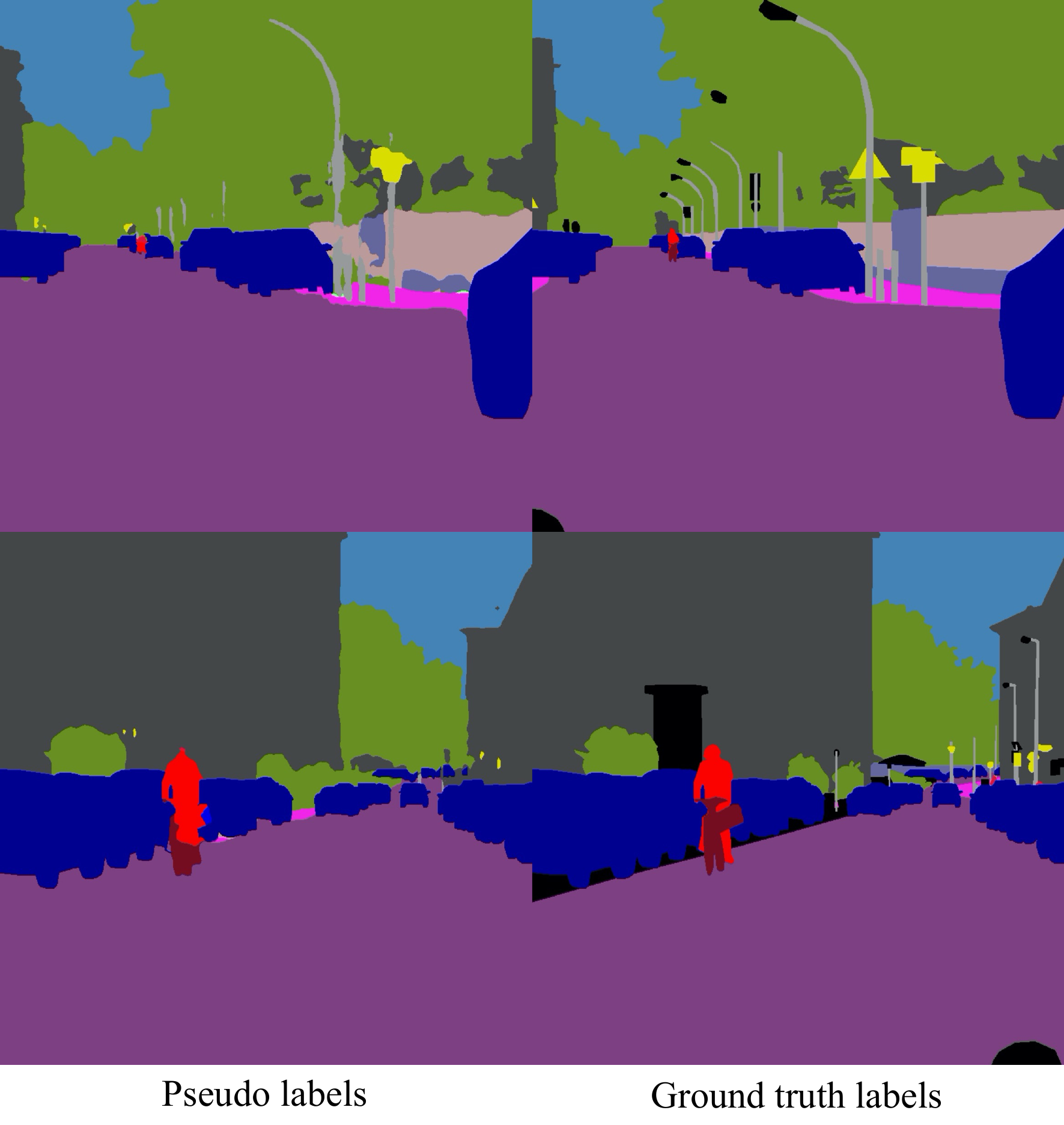}
  \vspace{-0.1cm}
  \caption{\textbf{Pseudo labels and ground truth labels}. The labels of the novel class \textit{car} are the same, while in other places the ground truth labels provide more precise details.}
  \label{fig:pseudo}
  \end{center}
\vspace{-0.5cm}
\end{figure}

In the pseudo label method (PLM), the old trained final branch heads provide the prediction of old classes, and these predictions combine with the annotation of the new class to generate the pseudo label of the training sample. In this way, labelers only need to give annotations for the new class and do not need to annotate for every pixel of the training samples. To verify that the PLM is reasonable, we conduct experiments using the ground truth labels for incremental learning rather than the pseudo labels. The results are in Table~\ref{tab:gt}. Compared to Table~\ref{tab:incre}, we find that the performance of using pseudo labels is similar to the performance of using ground truth labels, demonstrating the effectiveness of our PLM method. Some visualization of pseudo labels is shown in Fig.~\ref{fig:pseudo}.

\begin{table}[h]
\small
\renewcommand\arraystretch{1.1}
\renewcommand\tabcolsep{3pt}
\begin{center}
\begin{tabular}{lccccc}
\hline
\color[HTML]{009901} \textbf{16+1 setting} & mIoU & mIoU$\boldsymbol {\mathrm{_{novel}}}$ & mIoU$\boldsymbol{\mathrm{_{old}}}$  & mIoU$\boldsymbol{\mathrm{_{harm}}}$ \\ \hline
{$Q=1$ }  & {58.9} & 60.2  & {58.8} & 59.5 \\
{$Q=5$ }  & 61.2 & {72.3}  & 60.5 & 65.9 \\
{$Q=100$ }& 70.2     & 85.8 & 69.2  & 76.6 \\
{$Q=1000$ }& 72.0     & 91.8 & 70.8  & 79.9 \\
{$Q=2975$ }& \textbf{72.0}     & \textbf{91.9} & \textbf{70.8}  & \textbf{80.0} \\
\hline
\end{tabular}
\end{center}
\caption{\textbf{Incremental learning results using the ground truth under various $Q$}. Compared to Table~\ref{tab:incre}, this table shows the incremental learning results using the ground truth labels are similar to the results using pseudo labels.}
\label{tab:gt}
\vspace{-0.5cm}
\end{table}

\end{document}